\documentclass[10pt,twocolumn,letterpaper]{article}

\usepackage{cvpr}              %

\usepackage{multirow}
\usepackage[table]{xcolor}
\usepackage{xcolor}
\definecolor{bestcolor}{RGB}{252, 229, 205}
\newcommand{\best}[1]{\cellcolor{bestcolor}{\textbf{#1}}}
\definecolor{secondcolor}{RGB}{243, 243, 243}
\newcommand{\second}[1]{\cellcolor{secondcolor}{\underline{#1}}}

\usepackage{pifont} %
\definecolor{darkergreen}{RGB}{21, 152, 56}
\definecolor{red2}{RGB}{252, 54, 65}
\newcommand{\yesmark}{\textcolor{darkergreen}{\ding{52}}}
\newcommand{\nomark}{\textcolor{red2}{\ding{56}}}

\definecolor{cvprblue}{rgb}{0.21,0.49,0.74}
\usepackage[pagebackref,breaklinks,colorlinks,citecolor=cvprblue]{hyperref}

\usepackage{xspace}

\newcommand{\rawrefimg}{\ensuremath{x_{i_R}}\xspace}
\newcommand{\rawcond}{\ensuremath{x_{c}}\xspace}

\newcommand{\rawtarimg}{\ensuremath{x_i}\xspace}
\newcommand{\rawtriplet}{\ensuremath{\langle \rawrefimg, \rawcond, \rawtarimg \rangle}\xspace}
\newcommand{\rawimgpair}{\ensuremath{\langle \rawrefimg, \rawtarimg \rangle}\xspace}

\newcommand{\oursfull}{Language Only training for Composed Image Retrieval\xspace}
\newcommand{\oursfullbold}{\textbf{L}anguage Only tra\textbf{in}ing for \textbf{C}omposed \textbf{I}mage \textbf{R}etrieval\xspace}
\newcommand{\ours}{LinCIR\xspace}
\newcommand{\smpfull}{Self-Masking Projection\xspace}
\newcommand{\smp}{SMP\xspace}

\newcommand{\sptkn}{\texttt{[\$]}}
\newcommand{\condtkn}{\texttt{[cond]}}

\newcommand{\aphotoprompt}{a photo of \sptkn}
\newcommand{\aphotopromptwithcond}{a photo of \sptkn that \condtkn}

\newcommand{\catprompt}{gray cat sleeps on a pillow}
\newcommand{\catpromptreplaced}{\sptkn sleeps on \sptkn}

\def\paperID{2080} %
\def\confName{CVPR}
\def\confYear{2024}

\title{Language-only Efficient Training of Zero-shot Composed Image Retrieval}

\author{Geonmo Gu$^{*,\,1}$ \quad Sanghyuk Chun$^{*,\,2}$ \quad Wonjae Kim$^{2}$ \quad Yoohoon Kang$^{1}$ \quad Sangdoo Yun$^{2}$\\
\\
{$^{1}${NAVER Vision} \qquad $^{2}${NAVER AI Lab}} \qquad {\small $^*$ Equal contribution}
}

\begin{document}
\maketitle

\begin{abstract}
Composed image retrieval (CIR) task takes a composed query of image and text, aiming to search relative images for both conditions. Conventional CIR approaches need a training dataset composed of triplets of query image, query text, and target image, which is very expensive to collect. Several recent works have worked on the zero-shot (ZS) CIR paradigm to tackle the issue without using pre-collected triplets. However, the existing ZS-CIR methods show limited backbone scalability and generalizability due to the lack of diversity of the input texts during training. We propose a novel CIR framework, only using language for its training. Our LinCIR (\textbf{L}anguage-only tra\textbf{in}ing for CIR) can be trained only with text datasets by a novel self-supervision named self-masking projection (SMP). We project the text latent embedding to the token embedding space and construct a new text by replacing the keyword tokens of the original text. Then, we let the new and original texts have the same latent embedding vector. With this simple strategy, LinCIR is surprisingly efficient and highly effective; LinCIR with CLIP ViT-G backbone is trained in 48 minutes and shows the best ZS-CIR performances on four different CIR benchmarks, CIRCO, GeneCIS, FashionIQ, and CIRR, even outperforming supervised method on FashionIQ.
Code is available at \href{https://github.com/navervision/lincir}{\texttt{github.com/navervision/lincir}}
\end{abstract}

\section{Introduction}

\begin{figure}
    \centering
    \includegraphics[width=.94\linewidth]{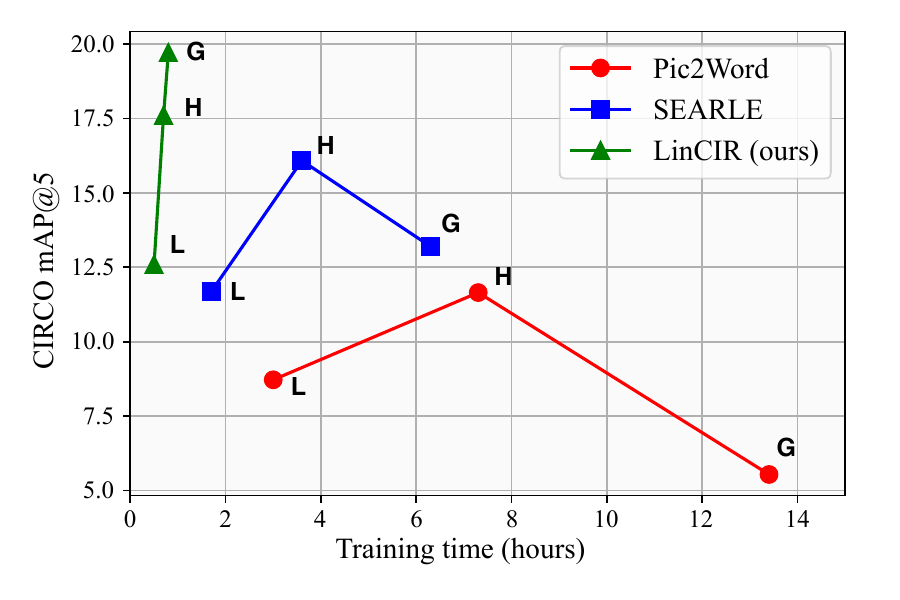}
    \vspace{-1.75em}
    \caption{\small {\bf Training time (hours) vs. Zero-shot Composed Image Retrieval (ZS-CIR) performance.} Thanks to our efficient language-only training strategy, our \ours outperforms the previous ZS-CIR methods in both training time and CIR performance. The training time is measured on 8 A100 GPUs. We compare the models on the CIRCO mAP@5 \cite{searle} score for a more comprehensive evaluation of CIR models (more results are in \cref{fig:tt_vs_perf}). Notably, when we scale up the backbone CLIP \cite{clip,openclip} model size by ViT-L, ViT-H and ViT-G, \ours shows a promising performance boost with surprisingly short training time (48 mins for ViT-G). On the other hand, Pic2Word \cite{pic2word} and SEARLE \cite{searle} cannot be scaled up to CLIP ViT-G due to their limitation on restricted textual expressions and the lack of diversity of input texts.}
    \label{fig:teaser}
    \vspace{-.85em}
 \end{figure}

Composed image retrieval (CIR) is a challenging vision-language (VL) task that takes a composed query of image and text, aiming to search relative images for both conditions \cite{cirr}.
As language serves as the most natural method for encoding human interaction, CIR provides a higher degree of freedom and a better user experience for image-based search engine applications, such as web commerce.

\begin{figure*}[t]
    \centering
    \includegraphics[width=.78\linewidth]{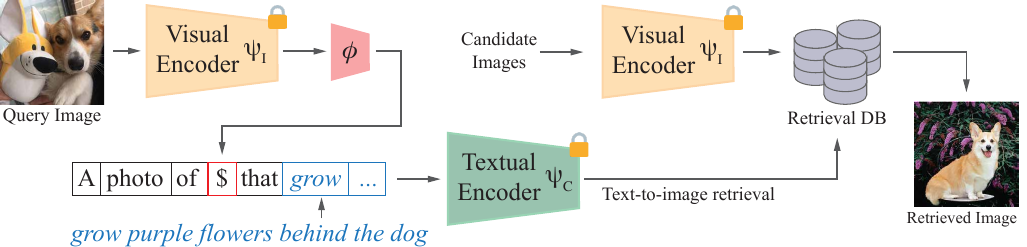}
    \vspace{-.5em}
    \caption{\small {\bf Overview of ZS-CIR with a projection to the token embedding space.} The mainstream ZS-CIR methods, such as Pic2Word \cite{pic2word}, SEARLE \cite{searle} and \ours (ours), train a projection module $\phi$ that projects the image latent embedding $z_i$ into the token embedding space $e_c$ with a custom prompt (\eg, \aphotopromptwithcond). The textual encoder output is used for CIR.}
    \label{fig:zs_cir}
    \vspace{-.5em}
\end{figure*}

One of the main challenges of CIR is the expensive dataset collection pipeline. CIR datasets consist of triplets \rawtriplet, where \rawrefimg is an image query, \rawcond is a text query, and \rawtarimg is the target image. Unlike image-text paired datasets, such as CC3M \cite{sharma2018conceptual_caption} or LAION \cite{schuhmann2022laion}, such triplets are almost impossible to collect by web crawling and require expensive human labor to create each triplet.
For example, the datasets are constructed by gathering candidates of \rawimgpair and manually annotating \rawcond by human annotators \cite{fashioniq, cirr}, which is hard to scalable. Therefore, the size of the training triplets is usually small (\eg, 46.6k \cite{fashioniq}, 28.8k \cite{cirr}), and the existing CIR methods trained on such triplets \cite{vo2019tirg,chen2020jvsm,chen2020val,jandial2020trace,dodds2020maaf,yu2020curlingnet,shin2021rtic,kim2021dcnet,liu2021cirplant,lee2021cosmo,jandial2022sac,tian2022aacl,chen2022mur,delmas2022artemis,baldrati2022clip4cir}
suffer from the lack of generalizability to diverse unseen domains.

\begin{figure*}[t]
    \centering
    \includegraphics[width=.93\linewidth]{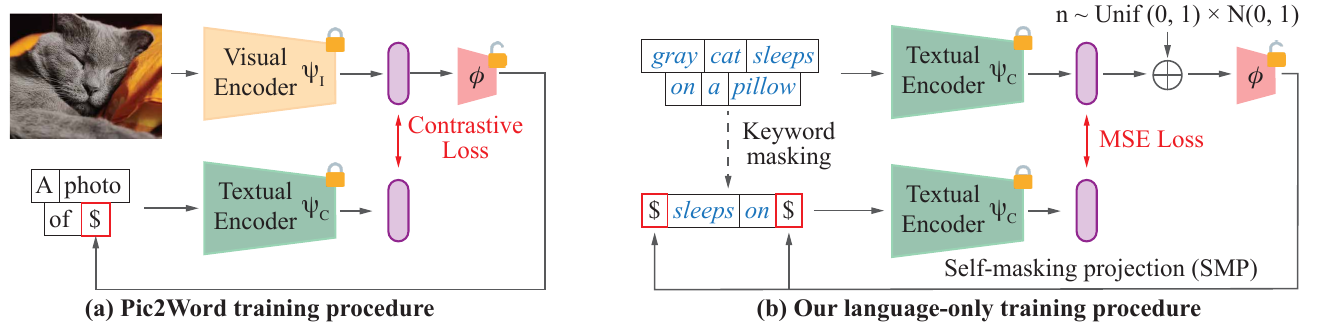}
    \vspace{-0.5em}
    \caption{\small {\bf Comparison of Pic2Word \cite{pic2word} and \ours training procedures.} (a) Pic2Word \cite{pic2word} and SEARLE \cite{searle} training procedure requires both the visual encoder and the textual encoder. They only need images for training, while the text prompt is pre-defined \cite{pic2word} or automatically generated \cite{searle}. (b) \ours is trained solely on texts with the frozen textual encoder. First, a projection module $\phi$ projects a textual latent embedding of a sentence $z_t$ into the token embedding space. Before the projection, a random noise $n$ is added to $z_t$ to reduce the modality gap between text and image. We introduce a new self-supervision, named \smpfull (\smp), by replacing all keywords of the given caption with the projected embedding by $\phi$ and extracting a modified text embedding $\widehat z_t$. Finally, the projection module $\phi$ is trained by the MSE loss between $z_t$ and $\widehat z_t$. Note that both (a) and (b) use the same inference strategy shown in \cref{fig:zs_cir}.}
    \label{fig:method_overview}
    \vspace{-.5em}
\end{figure*}

To overcome the drawback, recent studies have explored zero-shot CIR (ZS-CIR), a scalable direction, by eliminating the dependency on the pre-collected triplet datasets. For example, \citet{pic2word} and \citet{searle} propose projection-based ZS-CIR methods without using triplet datasets.
Based on the pre-trained CLIP \cite{clip}, they train a lightweight projection module $\phi$ that projects the CLIP image latent embedding $z_i$ to the CLIP text token embedding space $e_c$ (see \cref{fig:zs_cir}).
These approaches have shown promising generalizability to unseen datasets. However, these methods struggle to handle diverse text conditions because they rely on pre-defined naïve text prompts during training (\eg, \aphotoprompt). Moreover, their training frameworks need sequential forward operations of the visual and textual encoders (see \cref{fig:method_overview} (a)), resulting in inefficient training and limited scalability to a larger backbone.

In this paper, we introduce a new paradigm of ZS-CIR, named \oursfullbold (\ours)\footnote{Pronounced as ``linker'', meaning for linking the two modalities.}. As shown in \cref{fig:method_overview} (b), instead of projecting the image latent embedding $z_i$, we propose to project the text latent embedding $z_c$ to the token embedding space $e_c$.
We introduce a novel self-supervision, named \smpfull (\smp), for language-only training. We replace all the ``keywords'' of the original text with the projected text embedding of the original text to produce an embedding $\widehat z_c$ and apply MSE loss between $z_c$ and $\widehat z_c$. Here, we define ``keyword'' as consecutive adjectives and nouns.
For example, the keywords of ``\catprompt'' are ``gray cat'' and ``a pillow''; therefore, it becomes ``\catpromptreplaced''.
The purpose of \smp is to make \sptkn token interpreted as the ``one-word summarization'' of the input by extracting the essential information of the input.
During inference, we simply perform text-to-image retrieval by projecting image embedding to the token embedding space using the projection module $\phi$ as shown in \cref{fig:zs_cir}.
However, this strategy can suffer from the modality gap between textual and visual modalities \cite{liang2022mind}, \ie, even though our $\phi$ module works perfectly for text latent embeddings, it can underperform for the target visual embeddings. We mitigate the issue by employing a random noise addition strategy \cite{nukrai2022capdec}, carefully choosing a probability distribution that ensures the diversity of the noise-augmented textual embeddings.

Our paradigm has three advantages over the previous approaches.
First, while Pic2Word and SEARLE projection modules are trained with a restricted text prompt, \ie, \aphotoprompt, the projection module of \ours is trained with the diverse text inputs from the actual texts. Due to this reason, Pic2Word and SEARLE show degenerated performances when the backbone size becomes larger (see \cref{fig:teaser}). On the other hand, \ours is more generalizable to complex and diverse text conditions, showing superior ZS-CIR performances than others, especially for larger backbones.
Second, as \ours only utilizes the textual encoder, our training process is highly efficient and scalable than the methods incorporated with the visual encoder; our language-only training strategy is $\times$6.0 faster than Pic2Word \cite{pic2word} and  $\times$8.4 faster than SEARLE \cite{searle} with CLIP ViT-L backbone. When we scale up the backbone size to CLIP ViT-G, the gap becomes $\times$16.4 and $\times$17.6, respectively.
Even ViT-G training for \ours only takes 48 minutes using 8 A100 and less than 2 hours using 1 V100.
Third, our method is storage-efficient; for example, the CC3M dataset \cite{sharma2018conceptual_caption} images occupy about 430GB storage size, while its captions only need 125MB.
We train \ours only with 571MB storage size for storing the 5.5M training captions.
In summary, \ours shows the best training time and ZS-CIR performance as shown in \cref{fig:teaser} and \ref{fig:tt_vs_perf}.

Our contribution can be summarized as follows: (1) We propose \ours, a novel and efficient language-only training framework for ZS-CIR. (2) We introduce a new self-supervision for language-only training, named \smpfull (\smp). (3) We employ a better random noise addition strategy than naïve Gaussian noise to mitigate the modality gap. (4) \ours achieves the best training time and the ZS-CIR performances on four ZS-CIR benchmarks (CIRCO \cite{searle}, GeneCIS \cite{genecis}, FashionIQ \cite{fashioniq} and CIRR \cite{cirr}). Notably, \ours even outperforms the state-of-the-art supervised method \cite{baldrati2022clip4cir} on FashionIQ.

\section{Preliminaries}

\paragraph{Vision-langauge models (VLM).}
As with previous ZS-CIR methods, \ours utilizes a pre-trained VLM, such as CLIP \cite{clip, openclip} or BLIP \cite{blip}. We use VLMs that map an image input $x_i$ and a text input $x_c$ to the $d$-dimensional joint embedding space.
Here, a given caption $x_c$ is tokenized by the pre-defined tokenizer as $t_c = \{t_c^k ~|~ k=1 \cdots K \}$, where $K$ is the number of tokens, and mapped to \emph{token embeddings} $e_c = E_w(t_c) = \{E_w(t_c^k) ~|~ k = 1 \cdots K \}$, where $E_w$ is the embedding layer parameterized by $w$. Afterwords, a textual encoder $\psi_C$ encodes the token embeddings to extract \emph{textual latent embeddings} $z_c = \psi_C(x_c) = \psi_C(E_w(t_c))$.
The \emph{visual latent embeddings} are extracted by the visual encoder $\psi_I$: $z_i = \psi_I(x_i)$. We will use the terminology \emph{token embeddings} to represent $e_c$, and use \emph{textual (or visual) latent embeddings} to represent $z_c$ (or $z_i$).

\vspace{-.5em}
\paragraph{VLM modality gap.}
The VLM joint embedding space suffers from the semantic gap between each modality. Namely, the visual and the textual latent embeddings are not exactly aligned with each other, but they are located in completely separate regions of the embedding space \cite{liang2022mind}.

Such a modality gap hinders the harmonization of visual and textual latents in the joint embedding space. To mitigate the gap, language-only training has been recently introduced for image captioning tasks \cite{nukrai2022capdec,li2023decap,gu2023can}. They train a text decoder from the CLIP textual embeddings to generate image captions from visual embeddings. During training the decoder, the modality gap is bridged by injecting Gaussian noises \cite{nukrai2022capdec,gu2023can} or projection-based method \cite{li2023decap}.

In this paper, we utilize language-only training and a noise-addition strategy to mitigate the modality gap. We carefully studied the impact of noise and found a better distribution rather than simple Gaussian noise. We believe our findings can be transferred to other language-only training methods. Moreover, our target task, CIR, needs to model a relationship between triplets of \rawtriplet, whereas previous language-only training methods only consider the pair-wise relationship (\ie, image captioning).  We tackle this problem by proposing a novel self-supervision, named \smpfull (\smp), which injects the projected textual embeddings into the original token embeddings.

\vspace{-.5em}
\paragraph{CIR by projection to token embeddings.}
The mainstream ZS-CIR methods, such as Pic2Word \cite{pic2word} and SEARLE \cite{searle}, employ a projection-based method. Namely, they learn a projection module $\phi$ from the image latent embedding space to the token embedding space. For inference, they project the image latent embedding $z_i$ to the token embedding (\texttt{\$}), then perform text-to-image retrieval with the prompt ``\aphotopromptwithcond'',
where \texttt{[cond]} is a text condition. 
\cref{fig:zs_cir} shows an overview of how textual projection-based ZS-CIR works. The main research point of this field is how to train the projection module $\phi$ to capture the visual information into the token embedding space.

Pic2Word \cite{pic2word} trains the projection module $\phi$ by minimizing contrastive loss between image latent embedding 
and the textual latent embedding of ``\aphotoprompt''
(see \cref{fig:method_overview} (a)). SEARLE \cite{searle} employs a similar approach to Pic2Word. First, they employ optimization-based textual inversion to generate pre-defined special tokens for an image and train $\phi$ to predict the token embedding. SEARLE employs CLIP zero-shot classification to predict the ``concept'' of the given image and refine the prompt by letting GPT \cite{brown2020gpt3} continue the phrase. Both methods only use image inputs $x_i$ for training without accessing the CIR triplets.

Although Pic2Word and SEARLE achieve reasonable ZS-CIR performances, they have two significant problems. First, they heavily rely on the initial prompt,
``\aphotoprompt'', limiting the diversity of the textual encoder input. As \cref{fig:teaser}, we argue that diversifying the input texts during training the projection module $\phi$ is critical to train with larger backbones (\eg, CLIP ViT-G), while using the naïve prompt is failed to scale up the backbone size. Second, they need image inputs, which are less compact and redundant than text datasets. Moreover, the visual encoder usually needs more computation resources than the textual encoder because the visual encoder takes the fixed length token (\eg, 256). In contrast, the textual encoder takes shorter token lengths (\eg, average token length of CC3M $\approx$10). We tackle the problems by introducing a language-only training method, showing remarkable efficiency and scalability.

\section{Language-only Training of Zero-shot CIR}
\label{sec:method}

This section introduces a new paradigm for ZS-CIR, named \oursfull (\ours). We first introduce a novel language-only self-supervision, \smpfull (\smp), that enables a language-only training for CIR (\S\ref{subsec:smp}). Then, we explain the modality gap problem and empirically show that adding a carefully chosen random noise can mitigate the problem (\S\ref{subsec:noise_design}). Finally, we describe the advantage of \ours in terms of efficiency and scalability (\S\ref{subsec:efficiency_scalability}).

\subsection{\smpfull (\smp)}
\label{subsec:smp}

We aim to learn a projection module $\phi$ that captures and retains the original visual information after the projection and the textual encoder. While the previous methods focus on directly mapping visual information to the token space with a naïve prompt (\ie,  \aphotoprompt), we argue that focusing on the textual encoder is more important. Our zero-shot CIR is based on text-to-image retrieval (see \cref{fig:zs_cir}). It means that the quality of the textual latent embedding is more critical to the final ZS-CIR performances. Hence, rather than focusing on minimizing the gap between visual information and the naïve prompt, we aim to achieve a projection module $\phi$ that captures the semantics of the keywords in the given text.

To achieve our goal, we introduce a novel language-only self-supervision named \smpfull (\smp). First, we project the textual embedding $z_c$ of a given text input $x_c$ with the projection module $\phi$ to the token embedding space, \ie, $\widehat e_c = \phi(z_c)$, where $\widehat e_c$ is the projected textual embedding. Then, we replace the token embeddings of all the ``keywords'' of $x_c$ with the projected token embedding $\widehat e_c$.
We define the keywords of the sentence as consecutive nouns and adjectives. For example, the keywords of ``A Russian Blue cat is gray and cute'' will be ``A Russian Blue cat'', ``gray'' and ``cute''; hence it will be converted to ``\sptkn is \sptkn and \sptkn'', where \sptkn is a special token to represent the projected token embedding $\widehat e_c$.
By treating all the main concepts (keywords) in the caption as the same \sptkn, we intend \sptkn to represent the overall essential information of the inputs. Note that similar to \texttt{[MASK]} tokens of masked modeling, \sptkn with different positions will be encoded in different features due to the positional embeddings.
Using the converted caption, we extract a converted textual latent feature $\widehat z_c$ and minimize MSE loss between the original textual embedding $z_c$ and $\widehat z_c$. Note that we only train the $\phi$ module while keeping the textual encoder frozen.

The intuition behind \smp is that semantic information is not balanced across the tokens, but concentrated on the specific keywords. We assume that it is more common that adjectives and nouns in the sentence are more important than other part-of-speechs (POS), such as verbs or adverbs. We empirically observe that our design choice (replacing all keyword token embeddings with $\widehat e_c$) is the best among the other variants, such as randomly replacing $n$ keyword tokens ($n=1, 3, 5$), replacing a random token, replacing all non-keyword tokens, or replacing all noun tokens (See \cref{tab:abl_masking}).

\smp has two benefits over the previous image-based ZS-CIR supervision \cite{pic2word,searle}. First, \smp allows the textual encoder to accept more diverse captions rather than ``\aphotoprompt''. While previous methods risk being sensitive to natural sentence variations, potentially affecting performances, our approach replaces tokens in natural sentences, maintaining robust performance across a more diverse set of sentence constructions.
Second, \smp only requires language inputs; therefore, the overall training procedure is efficient regarding the training time and the storage size. It means that \ours can easily scale up in terms of the backbone size and the dataset scale. We will discuss the efficiency and the scalability of \ours in the \cref{subsec:efficiency_scalability}.

\begin{figure*}[t]
     \centering
     \begin{subfigure}[b]{0.248\textwidth}
         \centering
         \includegraphics[width=\textwidth]{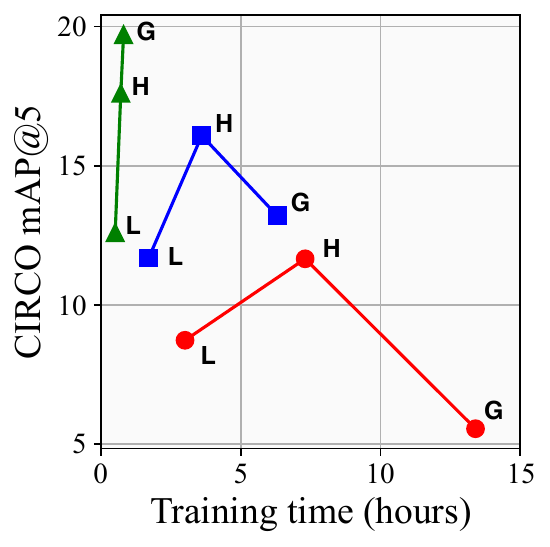}
     \end{subfigure}%
     \hfill
     \begin{subfigure}[b]{0.248\textwidth}
         \centering
         \includegraphics[width=\textwidth]{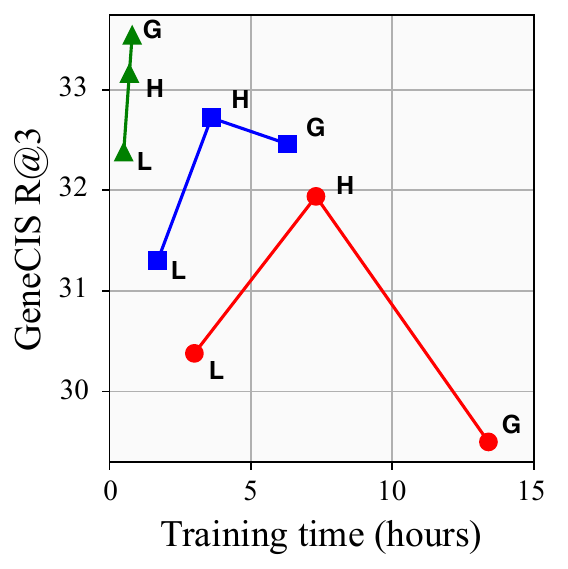}
     \end{subfigure}%
     \hfill
     \begin{subfigure}[b]{0.248\textwidth}
         \centering
         \includegraphics[width=\textwidth]{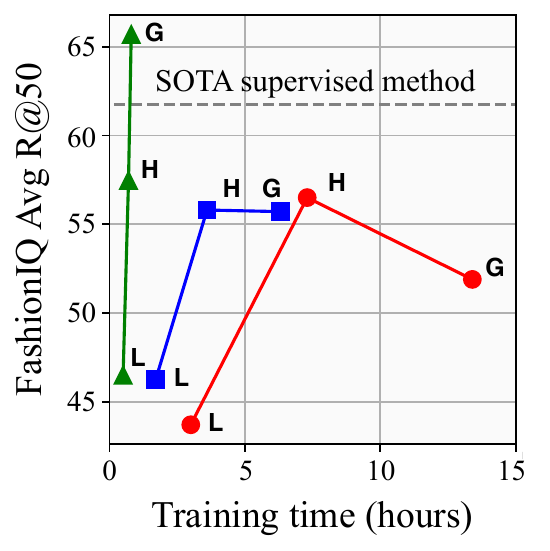}
     \end{subfigure}%
     \hfill
     \begin{subfigure}[b]{0.256\textwidth}
         \centering
         \includegraphics[width=\textwidth]{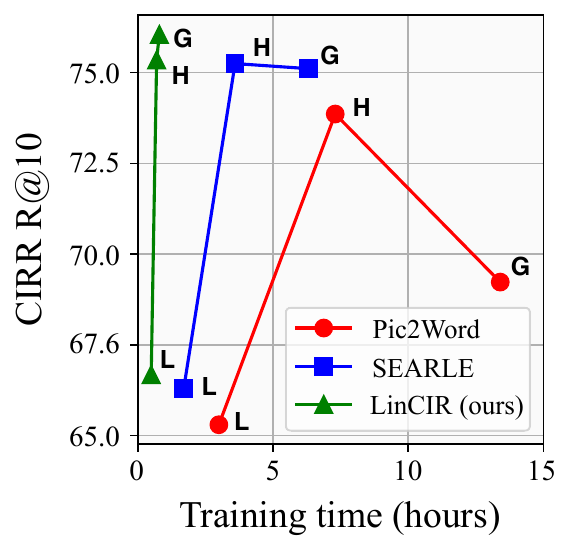}
     \end{subfigure}%
     \vspace{-1em}
     \caption{\small {\bf Training time vs. CIR performances.} We evaluate three CIR methods with three backbone sizes: ViT-L, ViT-H, and ViT-G. To avoid an unreliable assessment due to the nature of R@1, CIR performances are measured in CIRCO mAP@5 \cite{searle}, GeneCIS average R@3 \cite{genecis}, FashionIQ Average R@50 \cite{fashioniq}, and CIRR average R@10 \cite{cirr}. In all evaluation results, \ours achieves the best training time-performance trade-off. Moreover, Pic2Word and SEARLE show degenerated performances when scaling up the backbone size.}
     \label{fig:tt_vs_perf}
     \vspace{-.5em}
\end{figure*}

\subsection{Searching for a better noise distribution for reducing the modality gap.}
\label{subsec:noise_design}

\begin{table}[t]
\small
\setlength{\tabcolsep}{3pt}
\centering
\resizebox{\columnwidth}{!}{
\begin{tabular}{@{}llllllll@{}}
\toprule
& No noise & Student-t & Exp & $\chi^2$ & $\mathcal N(0, 1)$ & Unif(-1,1) & Ours \\ \midrule
L & 0.81 {\tiny (19.8)} & 0.76 {\tiny (23.1)} & 0.76 {\tiny (23.5)} & 0.78 {\tiny (23.5)} & 0.77 {\tiny (23.7)} & 0.74 {\tiny (25.1)} & \textbf{0.71 {\tiny (25.5)}} \\
H & 0.63 {\tiny (31.8)} & 0.59 {\tiny (32.4)} & 0.67 {\tiny (28.3)} & 0.64 {\tiny (27.0)} & 0.60 {\tiny (32.8)} & 0.59 {\tiny (33.9)} & \textbf{0.53 {\tiny (34.8)}} \\
G & 0.51 {\tiny (33.3)} & 0.55 {\tiny (36.1)} & 0.63 {\tiny (30.8)} & 0.56 {\tiny (30.7)} & 0.55 {\tiny (35.9)} & 0.58 {\tiny (35.3)} & \textbf{0.48 {\tiny (36.9)}} \\
\bottomrule
\end{tabular}
}
\caption{\small {\bf Modality gap vs. distributions.} Modality gap \cite{liang2022mind} (lower denotes less gap) on CC3M and CIRR dev R@1 (higher denotes better performance -- in the parentheses) for different noises with different backbone sizes (from ViT-L/14 to ViT-G/14).}
\label{tab:reb_gap}
\end{table}

Although \smp enables the language-only training, we still suffer from the modality gap between textual and visual embedding space \cite{liang2022mind}. Namely, even if the $\phi$ module works perfectly for language inputs, it can fail to be generalized to visual inputs. To tackle the problem, we employ a simple noise addition strategy following \citet{nukrai2022capdec}: we add a random noise before the projection during training.
\citet{nukrai2022capdec} employed a simple Gaussian noise, but we empirically observe that Gaussian noise is not effective in mitigating the gap. \cref{tab:reb_gap} shows the modality gap \cite{liang2022mind} measured by various CLIP backbones by adding different probability distributions to textual embeddings. In the table, we observe that the careful choice of distribution greatly affects the modality gap and the final performances.

We also observe that the generally used probabilistic distributions can suffer from a curse of dimensionality in the CLIP embedding space dimensions (\eg, 768-dim). The norm histogram of each probabilistic distribution (\cref{fig:norm_dist}) shows that the samples drawn from a Gaussian distribution have almost identical norm sizes.
From this observation, we employ a probability distribution enforcing the diverse norm sizes instead of the Gaussian distribution. We multiply a random scalar value by a random vector drawn from a Gaussian distribution, \ie, $n \sim \text{Unif}(0, 1) \times \mathcal N(0, 1)$. 
Conceptually, our distribution is a randomized Gaussian distribution with varying variances. In the Appendix, we illustrate that our design choice shows a more diverse norm distribution than the other distributions (See \cref{fig:norm_dist}).

In our experiments, we empirically observe that the noise addition strategy improves the overall CIR performances with a large gap by bridging the modality gap. We also empirically observe that our design choice outperforms other probability distributions with a large gap by diversifying the impact of the random noise (see \cref{tab:abl_noise}).

\subsection{Efficiency and scalability}
\label{subsec:efficiency_scalability}

\begin{table}
\small
\centering
\setlength{\tabcolsep}{1.6pt}
\begin{tabular}{lcccccccc}
\toprule
     & \multicolumn{4}{c}{Visual encoder} & \multicolumn{3}{c}{Textual encoder} \\
     & \# layers & $d$ & \# tokens & TP & \# layers & $d$ & \# tokens & TP \\ \midrule
ViT-L& 24 & 1024 & 256 & 18.5 & 12 & 768 & 12.56 & 65.5 \\
ViT-H& 32 & 1280 & 256 & 16.3 & 24 & 1024 & 12.56 & 35.5 \\
ViT-G& 48 & 1664 & 256 & 11.8 & 32 & 1280 & 12.56 & 26.5 \\ \bottomrule
\end{tabular}
\caption{\small {\bf Configuration of CLIP visual and textual encoders.} Every visual encoder uses the patch size 14 and the input resolution 224$\times$224. The text token length is the average token length of the CC3M \cite{sharma2018conceptual_caption} captions by the CLIP tokenizer. $d$ denotes the hidden dimension of Transformer blocks, and TP denotes throughput per second (higher means faster). TPs are measured by 1 V100 on CC3M \cite{sharma2018conceptual_caption} captions and COCO \cite{lin2014coco} images with FP16 weights.}
\label{tab:model_config}
\end{table}

The most remarkable advantages of \ours beyond its generalizability are the training efficiency and scalability. First, a text dataset is storage-efficient; the caption storage size of CC3M dataset \cite{sharma2018conceptual_caption} is only 125 MB, while its image storage size is about 430GB, about 3,400 times larger. Second, the forward complexity of the textual encoder is notably lower than that of the visual encoder.
For example, as shown in \cref{tab:model_config}, the textual encoder has fewer depth, dimension size, and input token size than the visual encoder taking 224 $\times$ 224 resolution images.
As a result, the average inference time of the CLIP ViT-L visual encoder is $\times$3.5 times slower than that of the textual encoder (\cref{tab:model_config}). Furthermore, the average throughput of the ViT-G textual encoder is even $\times$1.4 times faster than the ViT-L visual encoder.

All these advantages make \ours easily scalable. 
Even though we increase the backbone size, the overhead of the textual encoder is not significantly increased.
We can train \ours with the CLIP ViT-G backbone in 48 minutes using 8 A100 GPUs and 2 hours using a single V100.

\section{Experiments}

\subsection{Implementation details}

We use three-layered MLP for the $\phi$ model: \texttt{LN} \cite{layernorm} - \texttt{Linear} - \texttt{GeLU} \cite{gelu} - \texttt{Linear} - \texttt{GeLU} - \texttt{Linear} - \texttt{LN}. The intermediate hidden dimension is set to $4d$ ($d$ for each architecture is shown in \cref{tab:model_config}). We do not apply the $\ell_2$-normalization to the textual encoder outputs during training because, as shown in \cref{fig:norm_dist}, the added random noises have larger norm sizes than 1. If we apply the $\ell_2$-normalization, we observe that the $\phi$ module is not converged. Keywords of the given text are extracted by the POS tagger of \texttt{spacy} library. We use the AdamW optimizer \cite{adamw} with a fixed learning rate of 0.0001, weight decay of 0.01, and mini-batch size of 512. Dropout with probability 50\% is applied for the regularization.
We use CC3M \cite{sharma2018conceptual_caption} captions and 2.47M number of the curated StableDiffusion prompts\footnote{\url{https://huggingface.co/datasets/FredZhang7/stable-diffusion-prompts-2.47M}} for the training dataset (\ie, there are 5.5M training captions).

For a fair comparison between models, we select the model showing the best zero-shot CIRR \cite{cirr} dev R@1 score for the model selection. We employ an early stopping strategy by monitoring the validation score. We evaluate the CIR performances of the selected model in a zero-shot manner, \ie, one model is evaluated on four benchmarks.
We employ the visual and textual encoders of the official CLIP ViT-L \cite{clip}, and OpenCLIP ViT-H and ViT-G \cite{openclip}. In the Appendix, we show that \ours can be easily extended to the other VLMs, such as BLIP \cite{blip}.

\subsection{Experimental protocols}

\paragraph{Evaluation benchmarks and metrics.}
As pointed out by \citet{searle}, the existing CIR benchmarks only have a single positive, which can cause an unreliable evaluation. A similar phenomenon is also reported in the image-text cross-modal retrieval problem by \citet{eccv_caption}; such benchmarks can lead to a wrong model comparison result. For this reason, we use CIRCO as the main benchmark, which has multiple positives and measures a more reliable ranking-based metric, mAP@K \cite{musgrave2020metric}. We also report the R@K evaluation results on three additional datasets, GeneCIS \cite{genecis}, FashionIQ \cite{fashioniq}, and CIRR \cite{cirr}. We describe the details of each dataset in the Appendix.

In this paper, we argue that R@1 results can be somewhat noisy due to the false negatives in the dataset. Note that these benchmarks only have a unique positive triplet for each query \rawtriplet, \ie, if other plausible images (\ie false negatives) exist in the gallery set, the R@1 score in these benchmarks cannot correctly measure the actual retrieval performance. Due to this reason, we will focus on the mAP score if it is available. Otherwise, we will concentrate on R@K with a larger K (\eg, 10) rather than R@1.

\vspace{-.5em}
\paragraph{Comparison methods.}
We compare \ours with the recent ZS-CIR methods: Pic2Word \cite{pic2word} and SEARLE \cite{searle}. For a fair comparison, we train all methods with the same backbone architecture as \ours, namely ViT-H and ViT-G CLIP backbones. ViT-L results are measured using the official checkpoints.
We did not directly compare our method with recent methods that require massive external triplet datasets or take a long training time \cite{compodiff,ventura23covr}. More comparisons with these methods can be found in the Appendix.

\begin{table}[t]
\small
\setlength{\tabcolsep}{3pt}
\begin{tabular}{llcccc}
\toprule
 && mAP@5 & mAP@10 & mAP@25 & mAP@50 \\ \midrule
\multirow{3}{*}{ViT-L} & Pic2Word$^\dagger$ & 8.72 & 9.51 & 10.64 & 11.29 \\
&SEARLE$^\dagger$ & 11.68 & 12.73 & 14.33 & 15.12 \\
&\ours & \best{12.59} & \best{13.58} & \best{15.00} & \best{15.85} \\ \midrule
\multirow{3}{*}{ViT-H} & Pic2Word & 11.65 & 12.33 & 13.71 & 14.43 \\
&SEARLE & 16.08 & 16.92 & 18.81 & 19.69 \\
&\ours & \best{17.60} & \best{18.52} & \best{20.46} & \best{21.39} \\ \midrule
\multirow{3}{*}{ViT-G} & Pic2Word & 5.54 & 5.59 & 6.68 & 7.12 \\
&SEARLE & 13.20 & 13.85 & 15.32 & 16.04 \\
&\ours & \best{19.71} & \best{21.01} & \best{23.13} & \best{24.18} \\ \bottomrule
\end{tabular}
\caption{\small {\bf CIRCO results.} Results of Pic2Word \cite{pic2word}, SEARLE \cite{searle}, \ours by using different CLIP backbones are shown. $^\dagger$ denotes that the numbers are measured by the official checkpoint.}
\label{tab:main_circo}
\end{table}

\begin{table}[t]
\small
\centering
\begin{tabular}{llccc}
\toprule
 &  & R@1 & R@2 & R@3 \\ \midrule
\multirow{3}{*}{ViT-L} & Pic2Word$^\dagger$ & 11.16 & 21.47 & 30.38 \\
 & SEARLE$^\dagger$ & \best{12.26} & 22.11 & 31.30 \\
 & \ours & 12.19 & \best{22.76} & \best{32.38} \\ \midrule
\multirow{3}{*}{ViT-H} & Pic2Word & 11.89 & 22.17 & 31.94 \\
 & SEARLE & 13.34 & 23.72 & 32.72 \\
 & \ours & \best{13.76} & \best{23.87} & \best{33.16} \\ \midrule
\multirow{3}{*}{ViT-G} & Pic2Word & 10.67 & 20.70 & 29.50 \\
 & SEARLE & 12.87 & 22.61 & 32.46 \\
 & \ours & \best{13.66} & \best{24.64} & \best{33.54} \\ \bottomrule
\end{tabular}
\caption{\small {\bf GeneCIS results.} The average R@1, R@2, R@3 for ``Focus Attribute'', ``Change Attribute'', ``Focus Object'', and ``Change Object'' are shown. The full table is in the Appendix.}
\label{tab:main_genecis}
\vspace{-.5em}
\end{table}

\subsection{Main results}
\label{subsec:main_results}

The experimental results are summarized in \cref{fig:tt_vs_perf}: \ours outperforms the comparison methods in training time and retrieval performances. In all benchmarks, we observe that while the performance of \ours is enhanced by enlarging the backbone size, Pic2Word and SEARLE show inferior performances with the ViT-G backbone. We presume that it is because Pic2Word and SEARLE have a limited understanding of complex text queries 
because their $\phi$ module are trained on texts not diverse enough (\ie, \aphotoprompt).
On the other hand, our $\phi$ module shows a better understanding of complex texts as \ours is trained on diverse real-world texts from the caption datasets.

\begin{table*}[t]
\small
\centering
\begin{tabular}{llccccccccccc}
\toprule
 &  & \multicolumn{2}{c}{Shirt} &  & \multicolumn{2}{c}{Dress} &  & \multicolumn{2}{c}{Toptee} &  & \multicolumn{2}{c}{Average} \\
 &  & R@10 & R@50 &  & R@10 & R@50 &  & R@10 & R@50 &  & R@10 & R@50 \\ \midrule
\multirow{3}{*}{ViT-L} & Pic2Word$^\dagger$ & 26.20 & 43.60 &  & 20.00 & 40.20 &  & 27.90 & 47.40 &  & 24.70 & 43.70 \\
 & SEARLE$^\dagger$ & 26.89 & 45.58 &  & 20.48 & 43.13 &  & 29.32 & 49.97 &  & 25.56 & 46.23 \\
 & \ours & \best{29.10} & \best{46.81} &  & \best{20.92} & \best{42.44} &  & \best{28.81} & \best{50.18} &  & \best{26.28} & \best{46.49} \\ \midrule
\multirow{3}{*}{ViT-H} & Pic2Word & \best{36.90} & 55.99 &  & 28.01 & 51.51 &  & 40.18 & 62.01 &  & 35.03 & 56.50 \\
 & SEARLE & 36.46 & 55.45 &  & 28.46 & 51.07 &  & 38.81 & 60.89 &  & 34.57 & 55.80 \\
 & \ours & \best{36.90} & \best{57.75} &  & \best{29.80} & \best{52.11} &  & \best{42.07} & \best{62.52} &  & \best{36.26} & \best{57.46} \\ \midrule
\multirow{3}{*}{ViT-G} & Pic2Word & 33.17 & 50.39 &  & 25.43 & 47.65 &  & 35.24 & 57.62 &  & 31.28 & 51.89 \\
 & SEARLE & 36.46 & 55.35 &  & 28.16 & 50.32 &  & 39.83 & 61.45 &  & 34.81 & 55.71 \\
 & \ours & \best{46.76} & \best{65.11} &  & \best{38.08} & \best{60.88} &  & \best{50.48} & \best{71.09} &  & \best{45.11} & \best{65.69} \\ \midrule
 \multicolumn{2}{c}{Combiner (supervised) \cite{baldrati2022clip4cir}$^\dagger$} & 39.99 & 60.45 && 33.81 & 59.40 && 41.41 & 65.37 && 38.32 & 61.74 \\ 
 \bottomrule
\end{tabular}
\caption{\small {\bf FashionIQ results.} $^\dagger$ denotes that the numbers are from the original paper. \ours ViT-G even outperforms the previous state-of-the-art supervised CIR method \cite{baldrati2022clip4cir} with a large gap although \ours is not directly trained on the FashionIQ dataset.}
\label{tab:main_fiq}
\end{table*}

\begin{table}[t]
\small
\setlength{\tabcolsep}{3pt}
\centering
\begin{tabular}{llccccccc}
\toprule
 &  & \multicolumn{3}{c}{Full} && \multicolumn{3}{c}{Subset} \\
 &  & R@1 & R@5 & R@10 && R@1 & R@2 & R@3 \\ \midrule
\multirow{3}{*}{ViT-L} & Pic2Word$^\dagger$ & 23.90 & 51.70 & 65.30 && 53.76 & 74.46 & 87.08 \\
 & SEARLE$^\dagger$ & 24.24 & 52.48 & 66.29 && 53.76 & 75.01 & 88.19\\
 & \ours & {25.04} & {53.25} & \best{66.68} && {57.11} & {77.37} & {88.89}\\ \midrule
\multirow{3}{*}{ViT-H} & Pic2Word & 32.94 & 63.11 & 73.86 && 62.22 & 81.35 & 91.23\\
 & SEARLE & {34.00} & {63.98} & 75.25 && {64.63} & {83.21} & {92.77}\\
 & \ours & 33.83 & 63.52 & \best{75.35} && 62.43 & 81.47 & 92.12 \\ \midrule
\multirow{3}{*}{ViT-G} & Pic2Word & 30.41 & 58.12 & 69.23 && {68.92} & {85.45} & 93.04 \\
 & SEARLE & 34.80 & 64.07 & 75.11 && 68.72 & 84.70 & {93.23} \\
 & \ours & {35.25} & {64.72} & \best{76.05} && 63.35 & 82.22 & 91.98 \\ \bottomrule
\end{tabular}
\caption{\small {\bf CIRR results.} Due to the noisy nature of CIRR as a ZS-CIR benchmark, we only highlight the R@10 score for the full CIRR set. The detailed discussion can be found in \cref{subsec:main_results}}
\label{tab:main_cirr}
\vspace{-.5em}
\end{table}

We also provide the full evaluation results on the four benchmarks below. \cref{tab:main_circo} shows the evaluation results on the CIRCO dataset. In all experiments, \ours outperforms others with a significant gap. We can observe a similar finding in the GeneCIS average R@K results for four different subtasks (\cref{tab:main_genecis}), especially for R@K with a larger K. As shown in the Appendix, \ours outperforms other models, especially on ``Focus Attribute'' and ``Change Attribute'' tasks. We presume that it is because the Pic2Word and SEARLE training prompts are more specialized to objects (\aphotoprompt), where \ours can handle more detailed concepts in the image by altering all the nouns in the sentence (\eg, ``gray \texttt{[\$]} sleeps on a \texttt{[\$]}'').

In the FashionIQ benchmark (\cref{tab:main_fiq}), \ours even outperforms the state-of-the-art supervised method \cite{baldrati2022clip4cir} with a large gap (38.32 vs. 45.11 in the average R@10). Our work is the first ZS-CIR method that outperforms the supervised CIR method despite its generalizability to the other CIR benchmarks and flexibility for handling various conditions. 

We observe somewhat mixed results on CIRR (\cref{tab:main_cirr}): \ours achieves the best R@10 score, but not in some other metrics. We argue that this is because of two reasons. First, R@K with a small K cannot fully reflect the authentic performance. 
As shown in the Appendix, 
the retrieval results of \ours are plausible to humans, but because the dataset has incomplete positives (\ie, there are many false negatives), the R@1 score cannot correctly evaluate the model performance.
According to \citet{eccv_caption}, the similarity between the rankings measured by R@K on a partially annotated benchmark and those measured by mAP on a fully annotated benchmark becomes lower when we use a small K (\eg, 1).
Second, as observed by previous works \cite{pic2word,searle}, the quality of the CIRR benchmark as a ZS-CIR benchmark is somewhat doubtable. The CIRR text relative captions are often not truly relative (\ie, there exist false positives), and reference images can even be harmful to retrieval. As pointed out by \citet{searle}, this problem becomes more severe when we use a small subset of images, \ie, the subset R@Ks.
In summary, due to the noisy nature of CIRR full and subset R@1s, we propose to focus on the CIRR full R@10 scores rather than other metrics. In CIRR full R@10, \ours shows the same trends as the other benchmarks.

\subsection{Analysis}

\begin{table}[t]
\small
\centering
\begin{tabular}{lccc}
\toprule
           &  CIRR dev & \multicolumn{2}{c}{Fashion IQ} \\
Supervision design & R@1 & R@10 & R@50 \\ \midrule
\aphotoprompt \cite{pic2word} & 21.65 & 24.93 & 44.35 \\
Ours, but \sptkn extracted by $\psi_I$ & 22.63 & 22.01 & 39.87 \\
Our \smp design choice & \best{25.66} & \best{26.28} & \best{46.49} \\ \bottomrule
\end{tabular}
\caption{\small {\bf Impact of the supervision design.} Different target text designs (\eg, \catpromptreplaced in \cref{fig:method_overview} (b)) affect the performances. ``Ours, but \sptkn extracted by $\psi_I$'' denotes that the textual encoder before the $\phi$ module is replaced with the visual encoder.}
\label{tab:abl_supervision}
\end{table}
\begin{table}[t]
\small
\centering
\begin{tabular}{lccc}
\toprule
           &  CIRR dev & \multicolumn{2}{c}{Fashion IQ} \\
\smp Masking strategy & R@1 & R@10 & R@50 \\ \midrule
All non-keyword tokens & 20.59 & 19.61 & 36.97 \\
Random token & 22.98 & 22.44 & 40.92 \\
All noun tokens & 24.95 & 25.16 & 45.19 \\
\midrule
1 keyword token & 23.92 & 26.42 & 45.91 \\
3 keyword tokens  & 24.66 & \best{26.69} & \best{46.54} \\
5 keyword tokens & 25.14 & 26.28 & 46.29 \\ \midrule
All keyword tokens & \best{25.66} & {26.28} & {46.49} \\ \bottomrule
\end{tabular}
\caption{\small {\bf Impact of the \smp masking strategy.} The CIRR retrieval performances by varying the masking strategy for \smp (\ie, ``Keyword masking'' in \cref{fig:method_overview} (b)) are shown. We define ``keyword'' as consecutive adjectives and nouns, except for ``All noun tokens''. ``All noun tokens'' defines keywords as nouns.}
\label{tab:abl_masking}
\vspace{-.5em}
\end{table}

\begin{table}[t]
\small
\centering
\begin{tabular}{lccc}
\toprule
           &  CIRR dev & \multicolumn{2}{c}{Fashion IQ} \\
Noise type & R@1 & R@10 & R@50 \\ \midrule
No noise & 19.76 & 20.42 & 38.31 \\ \midrule
Student-t & 23.08 & 22.81 & 41.00 \\ 
Exponential & 23.51 & 25.78 & 45.26 \\
$\chi^2$ & 23.54 & 23.11 & 41.89 \\
$\mathcal N(0, 1)$ & 23.70 & 23.31 & 41.89 \\
Unif(-1, 1) & 25.14 & 25.88 & 45.78 \\ \midrule
$\mathcal N(0, 1) \times$ Unif(0, 1) & \best{25.47} & \best{26.05} & \best{46.29} \\ \bottomrule
\end{tabular}
\caption{\small {\bf Impact of the choice of the random noise.} Adding noise to the textual latent space helps to mitigate the inferior generalizability due to the modality gap. Moreover, using better random noise can significantly boost the overall performances.}
\label{tab:abl_noise}
\end{table}

\begin{table}[t]
\small
\setlength{\tabcolsep}{3pt}
\centering
\resizebox{\columnwidth}{!} {

\begin{tabular}{ccccccccc}
\toprule
&&&& CIRCO & GeneCIS & FashionIQ & CIRR & \multirow{2}{*}{Avg} \\
CC3M & SDP & COYO & OWT & mAP@5 & R@3 & R@10 & R@10 & \\ \midrule
\yesmark & \nomark & \nomark & \nomark & \best{13.72} & \second{32.80} & 25.11 & 64.95 & 34.15 \\
\nomark & \yesmark & \nomark & \nomark & 9.52 & 32.38 & 23.63 & 62.58 & 32.03 \\
\nomark & \nomark & \yesmark & \nomark & 11.36 & 31.48 & {26.18} & 65.33 & 33.59 \\
\nomark & \nomark & \nomark & \yesmark & 9.67 & 30.90 & 24.41 & 64.05 & 32.26 \\
\midrule
\yesmark & \yesmark & \nomark & \nomark & \second{12.59} & {32.38} & \second{26.28} & \second{66.68} & \best{34.48} \\ 
\yesmark & \yesmark & \nomark & \yesmark & 10.06 & \best{33.36} & 21.11 & 63.16 & 31.92 \\
\yesmark & \yesmark & \yesmark & \nomark & 11.54 & 32.08 & \best{26.97} & \best{66.80} & \second{34.35} \\
\bottomrule
\end{tabular}

}
\caption{\small {\bf Impact of the training corpus.}
OpenWebText (OWT) \cite{Gokaslan2019OpenWeb} and SD prompts (SDP) are text-only datasets and CC3M \cite{sharma2018conceptual_caption} and COYO-700M (COYO) \cite{kakaobrain2022coyo-700m} are image-text aligned datasets. In our experiments, we use CC3M + SDP for the training corpus, considering the dataset scale and overall CIR performances.}
\label{tab:abl_dataset}
\end{table}

In this subsection, we provide detailed analyses of our design choices.
If not specified, we compare the models on the CIRR dev split and FashionIQ test split.

\vspace{-.5em}
\paragraph{Impact of \smpfull (\smp) supervision.}
We compare two variants of \smp in \cref{tab:abl_supervision}. First, instead of employing our keyword masking strategy, we use ``\aphotoprompt'' as Pic2Word and SEARLE to train the $\phi$ module. Second, we replace the textual encoder before $\phi$ module with the visual encoder using the corresponding image of the caption. In the table, ``\aphotopromptwithcond'' variant performs worse than our design choice due to the limited diversity of the input texts. Interestingly, we observe that using both image and text pairs for \ours performs worse than our design choice. Note that the second model is trained on CC3M image-text pairs without using 2.47M SD prompts as our design choice. We presume it makes the $\phi$ model overfitted to CC3M image-text relationships, which significantly differ from our target CIR datasets.

\vspace{-.5em}
\paragraph{Impact of masking design choice.}
We compare other masking design choices, such as random tokens or non-keyword tokens, with our design choice in \cref{tab:abl_masking}. \cref{tab:abl_masking} shows that (1) masking the keywords performs better than masking the others. (2) Our keyword design -- \ie, consecutive adjectives and nouns -- is better than defining keywords as nouns. (3) The overall performances are enhanced by increasing the masked keywords. As the differences are not significant, we mask all the keywords for simplicity.

\vspace{-.5em}
\paragraph{Impact of the random noise addition.}
As we discussed in \cref{subsec:noise_design}, the random noise addition is critical to mitigating the modality gap. \cref{tab:abl_noise} supports this claim: when we do not add any noise, the performance becomes the worst. Our design choice shows the best performance among the other noise designs due to the diverse norm size of our distribution as shown in \cref{fig:norm_dist}.

\vspace{-.5em}
\paragraph{The impact of the training corpus.}
We evaluate the impact of the training corpus in \cref{tab:abl_dataset}.
We use four corpora, CC3M \cite{sharma2018conceptual_caption} (3M captions), StableDiffusion Prompts (SDP) (2.47M text prompts), COYO-700M (700M captions) \cite{kakaobrain2022coyo-700m} and OpenWebText (OWT) \cite{Gokaslan2019OpenWeb} (8M web texts). CC3M and COYO are image-text paired datasets, and OWT and SDP are text-only datasets.
The example training samples of each dataset are shown in the Appendix.
In \cref{tab:abl_dataset}, we observe that
models trained with image descriptions (CC3M and COYO) perform better than models trained with general texts because general web texts are often irrelevant to visual information. In addition, using multiple corpora improves the overall performance, \eg, CC3M (34.15) $\rightarrow$ CC3M + SDP (34.48). Although we observe that using more massive captions (\eg, 3M + 2.47M + 700M) can be helpful for some CIR tasks, such as FashionIQ and CIRR, we use CC3M and SDP for our training set considering the corpus size (3M + 2.47M), and the overall performances.

\paragraph{Qualitative results.}
We provide the additional qualitative retrieval results in the Appendix.
In summary, \ours shows qualitatively plausible retrieval performance even in a large-scale image database, \eg LAION-2B.
Also, we observe that the retrieved results by \ours often suffer from false negatives in benchmark datasets, showing the limitation of the R@1 evaluation on the existing CIR benchmarks.

\section{Conclusion}
We propose a novel zero-shot composed image retrieval (ZS-CIR) framework named \oursfull (\ours). \ours presents a breakthrough in addressing the challenges associated with the previous ZS-CIR methods. By leveraging a novel self-supervision technique, \smpfull (\smp), \ours eliminates the dependency on expensive CIR triplets, opting for a training process solely based on text inputs. This innovative approach significantly enhances scalability and generalizability, overcoming limitations observed in existing ZS-CIR methods. Notably, our \ours model achieves remarkable efficiency and ZS-CIR performances compared to other methods on multiple CIR benchmarks, including CIRCO, GeneCIS, FashionIQ, and CIRR. We underscore the effectiveness of our language-only training framework, offering a potent solution with wide-ranging implications for image retrieval tasks.

\onecolumn
\appendix
\numberwithin{equation}{section}
\numberwithin{figure}{section}
\numberwithin{table}{section}

\section{Dataset Details}

\subsection{CIR datasets}

\begin{figure}
\centering
\begin{subfigure}[b]{0.3\textwidth}
    \centering
    \includegraphics[width=.5\textwidth]{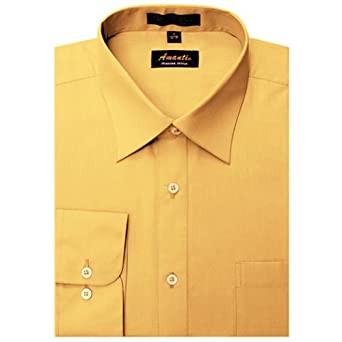}%
    \includegraphics[width=.5\textwidth]{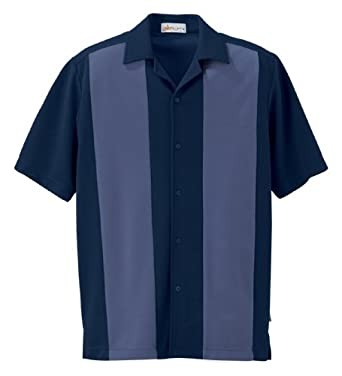}
    ``Is blue and has stripes''
    \caption{\bf FashionIQ}
    \label{fig:dataset_fiq}
\end{subfigure}
\hfill
\begin{subfigure}[b]{0.3\textwidth}
    \centering
    \includegraphics[width=.55\textwidth]{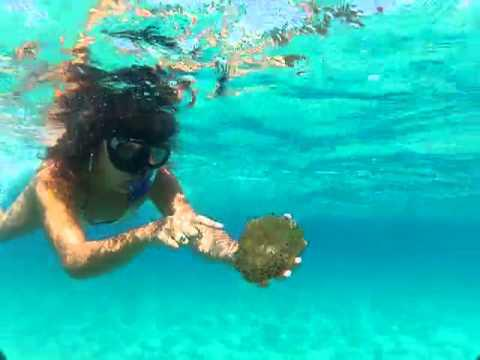}
    \hfill
    \includegraphics[width=.42\textwidth]{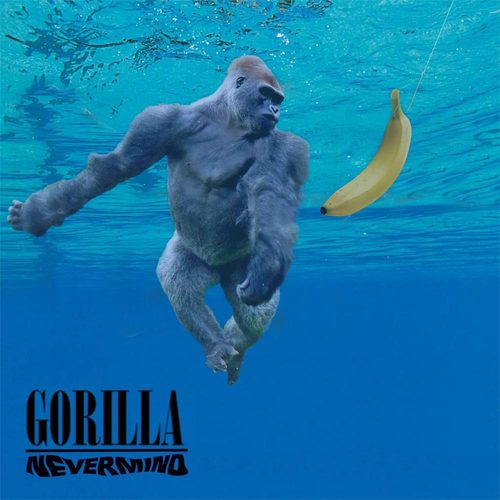}
    ``Same environment different species''
    \caption{\bf CIRR}
    \label{fig:dataset_cirr_b}
\end{subfigure}
\hfill
\begin{subfigure}[b]{0.3\textwidth}
    \centering
    \includegraphics[width=.38\textwidth]{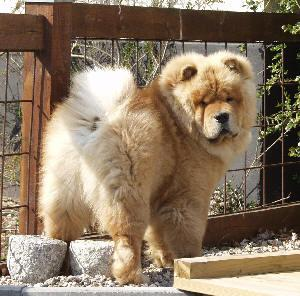}
    \hfill
    \includegraphics[width=.48\textwidth]{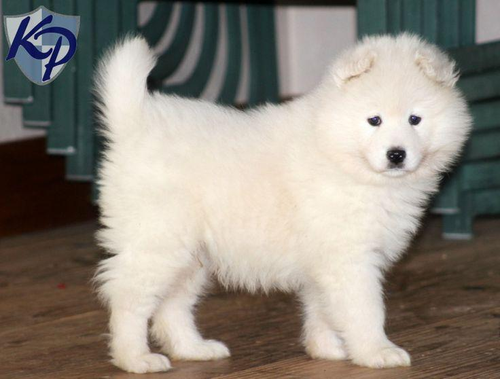}
    ``The target photo is of a lighter brown dog walking in white gravel along a wire and wooden fence''
    \caption{\bf CIRR}
    \label{fig:dataset_cirr_a}
\end{subfigure}
\hfill
\\
\begin{subfigure}[b]{\textwidth}
    \centering
    \includegraphics[width=.18\textwidth]{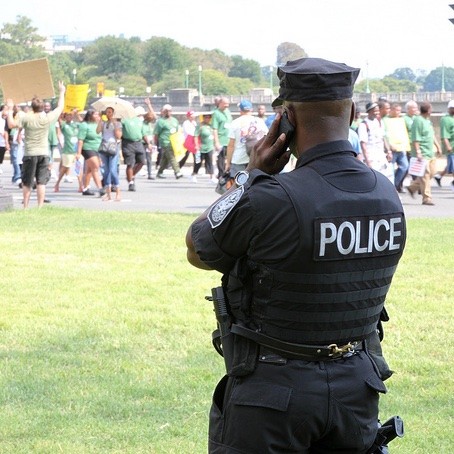}
    \hfill
    \includegraphics[width=.18\textwidth]{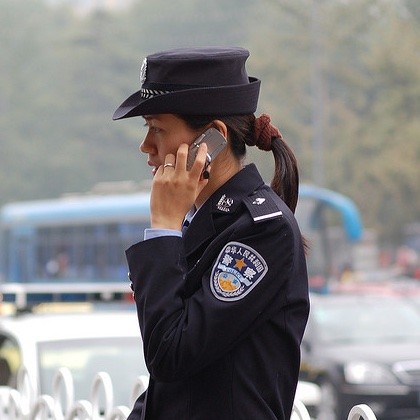}%
    \includegraphics[width=.18\textwidth]{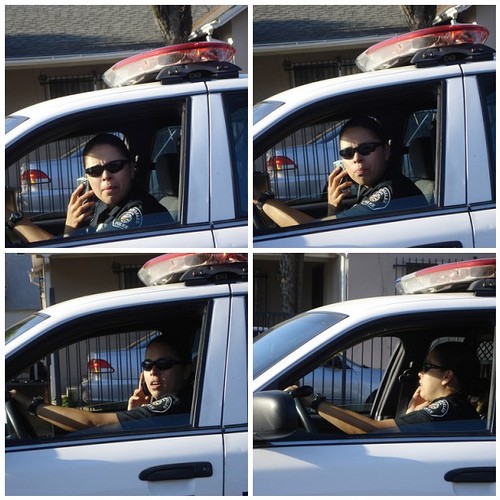}%
    \includegraphics[width=.18\textwidth]{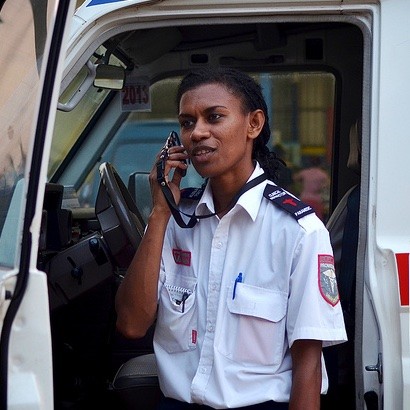}%
    \includegraphics[width=.18\textwidth]{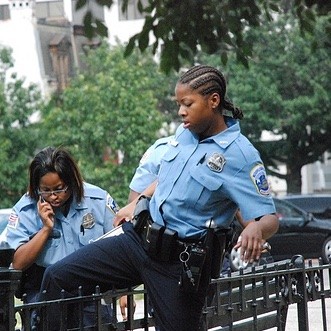}
    
    ``has a woman instead of a man and has a car in the background''
    \caption{\bf CIRCO}
    \label{fig:dataset_circo}
\end{subfigure}
\\
\begin{subfigure}[b]{0.4\textwidth}
    \centering
    \includegraphics[width=.48\textwidth]{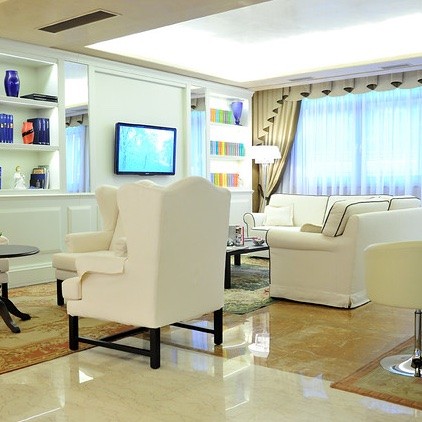}
    \hfill
    \includegraphics[width=.48\textwidth]{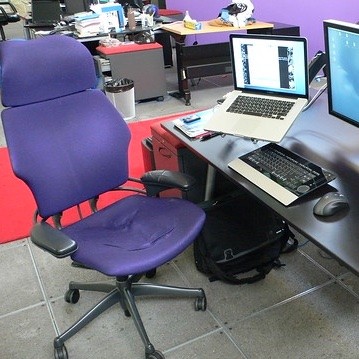}
    ``backpack''
    \caption{\bf GeneCIS ``Change Object''}
    \label{fig:dataset_genecis_change_obj}
\end{subfigure}
\hfill
\begin{subfigure}[b]{0.4\textwidth}
    \centering
    \includegraphics[width=.48\textwidth]{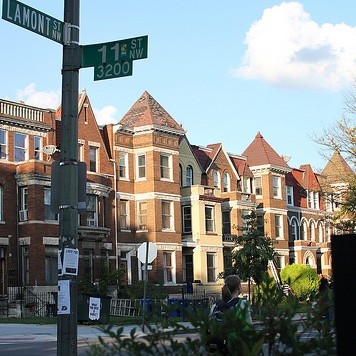}
    \hfill
    \includegraphics[width=.48\textwidth]{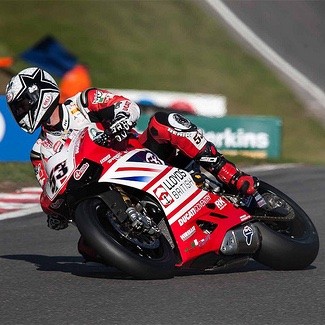}
    ``color''
    \caption{\bf GeneCIS ``Focus Attribute''}
    \label{fig:dataset_genecis_focus_att}
\end{subfigure}
\hfill
\caption{\small {\bf CIR Dataset examples.} In all examples, the first image is the reference, and the right image is the target image with the given caption. For CIRCO, the left image is the query image, and the other four images are all ground truth images with the given text query.}
\label{fig:dataset_examples}
\end{figure}

\noindent\textbf{FasionIQ \cite{fashioniq}} is a collection of fashion-related images in three categories: Shirt, Dress, and Toptee. FashionIQ contains 30,134 triplets from 77,684 images. During the dataset collection period, FashionIQ first collects the attributes of all images and then lets human annotators write a proper caption for highly relative images in terms of attributes. FashionIQ targets a realistic online shopping chat window. Therefore, FashionIQ collects captions using a visual chat-based interface, resulting in containing more realistic user text queries. The images are split into 6:2:2 for training, validation, and evaluation. As we aim to zero-shot CIR, we did not use the training split. Following the previous practice, we report the validation recalls because the labels of the evaluation split have not been publicly released. Examples of FashionIQ triplets are shown in \cref{fig:dataset_fiq}.

\vspace{.5em}
\noindent\textbf{CIRR \cite{cirr}} contains 21,552 real-life images sampled from NLVR$^2$ \cite{nlvr2}. CIRR also has training, validation, and test splits, where the test split is separately evaluated via the remote evaluation server. Therefore, we use the validation split of CIRR as the model selection criteria. While FashionIQ is limited to fashion-related domains, CIRR images are in more vast domains and have complex descriptions. During the dataset collection, CIRR collects visually similar images using ResNet-152 \cite{resnet} trained on ImageNet \cite{imagenet} and repeats the caption collection stage of FashionIQ. However, despite the realistic image domain, CIRR has two significant issues. First, while FashionIQ carefully collects the image pairs with human annotators, the pairs collected by CIRR are automatically collected with ImageNet-trained ResNet without careful human verification. It makes the pairs not actually visually similar and the paired images often significantly differ from each other. Second, while FashionIQ collects captions by letting the annotators mimic realistic online customers, CIRR lets the annotators write captions describing the differences between the images. Due to this reason, CIRR captions are unrealistic but contain unnecessary information or ambiguous descriptions, such as ``same environment different species'' (\cref{fig:dataset_cirr_b}) or ``The target photo is of a lighter brown dog walking in white gravel along a wire and wooden fence'' (\cref{fig:dataset_cirr_a}). For these reasons, CIRR is not realistic compared to FashionIQ. To resolve the issue, CIRR employs a retrieval task on a small subset, \eg, five items, but \citet{searle} observed that the subset retrieval task can be noisy because information of the target image is often not related to the reference image, but only related to the text condition. Furthermore, as \citet{searle}, \citet{pic2word} and \cref{fig:fn} observed, CIRR has a lot of false negatives (FNs) that can lead to wrong retrieval evaluation, as also shown in image-text cross-modal retrieval \cite{eccv_caption,pcmepp}.

\vspace{.5em}
\noindent Notably, both Fashion IQ and CIRR suffer from the FN problem; although the ground truth positive is one for each query, there could be multiple ground truths in the database. Furthermore, while FashionIQ collects triplets with careful verification, CIRR is constructed with noisy annotations. To tackle the issue, FashionIQ is evaluated by Recall@K with larger K (\eg, 10 or 50) and CIRR employs a small subset retrieval task. However, these approaches cannot resolve the problem fundamentally.

\vspace{.5em}
\noindent\textbf{CIRCO \cite{searle}} is based on COCO images \cite{lin2014coco} and contains multiple ground truths. CIRCO has 4.53 average ground truth images per query, enabling a more reliable and robust mAP metric \cite{musgrave2020metric,eccv_caption}. As CIRCO is designed for evaluating ZS-CIR methods, CIRCO has no training split. Instead, CIRO has a validation split (220 queries) and a test split (800 queries), where the test split is evaluated by the remote evaluation server. Example triplets are shown in \cref{fig:dataset_circo}.

\vspace{.5em}
\noindent\textbf{GeneCIS \cite{genecis}} consists of four conditional retrieval tasks: (1) focus on an attribute, (2) change an attribute, (3) focus on an object, and (4) change an object. GeneCIS focuses on defining the similarity in various notations, \eg, the similarity can be defined in the object ``with the same \textbf{car}'' or the attribute ``the same \textbf{color} as the car''. The attribute tasks are built upon VisualGenome \cite{visualgenome} with in-the-wild visual attributes from VAW \cite{vaw}. The object tasks are based on COCO \cite{lin2014coco} and its object classes. Each task has about 2,000 queries and the gallery size is only 15 (``Focus on an attribute'' task has 10) to avoid the FN problem, but to ensure only one positive for each query, similar to the CIRR subset strategy. Here, the text query is given by the name of the attribute or the object, \eg, ``backpack'' or ``color''. Example triplets are in \cref{fig:dataset_genecis_change_obj} and \cref{fig:dataset_genecis_focus_att}.

\vspace{.5em}
\noindent For all datasets, we use the same prompt for zero-shot composed image retrieval, namely, \aphotopromptwithcond where \condtkn is the given text condition. Example text conditions are shown in \cref{fig:dataset_examples}.

\subsection{Training corpora}

We provide examples of training corpora in
Tab. \textcolor{red}{10},
including CC3M \cite{sharma2018conceptual_caption}, StableDiffusion Probmpts (SDP)\footnote{\url{https://huggingface.co/datasets/FredZhang7/stable-diffusion-prompts-2.47M}}, OpenWebText \cite{Gokaslan2019OpenWeb}, and COYO-700M \cite{kakaobrain2022coyo-700m}. For OpenWebText, we only show one example of its shortened version, where the full document contains 1,386 words and 7,681 characters. We observe that the captions of CC3M are too generalized, therefore it has a weakness to describe an image in details. On the other hand, COYO-700M is too detailed, \eg, containing the exact product name, such as IPhone 6S. In practice, we use CC3M and SDP for representing a conceptualized caption for describing an image and detailed instruction for explaining an image by deep generative model, \ie, StableDiffusion.

\begin{itemize}
\item \textbf{CC3M:} ``person, was surprised by the staff'', ''red and white flag on the mast'', ``football player celebrates scoring for football team against football team in the final'', ``concept plug - in hybrid car on display'', ``a pencil drawing of a zebra and her baby.'', ``airline -- reasons why person leads the way in experience''
\item \textbf{SDP:} ``a full body character design by artgerm, greg rutkowski and alphonse mucha. sci - fi dagger. white tape and red translucent plastic tape project show attctive showgirl!! sci - fi helmet!! sharp edges. contour light effect!!. ultra detailed, elegant, intricate, octane render.'', ``realistic detailed face portrait of a beautiful young otherworldly ethereal alien geisha with blue hair by alphonse mucha, ayami kojima, yoshitaka amano, charlie bowater, karol bak, greg hildebrandt, jean delville, and mark brooks, art nouveau, neogothic, gothic, rich deep moody colors, celestial, surreal majestic winter pine dreamscape, character concept design'', ``cute anthropomorphic guinea pig full as an jedi in a spaceship, body portrait, divine lightning, by greg rutkowski, by charlie bowater'', ``boxing match between donald trump vs joe biden, stage lighting, award winning photo'', ``a chibi anime warrior with long red hair quickly swinging her sword in a full arc, dynamic slashing pose, detailed, anime'', ``saul goodman shaking hands with purple thanos at a walmart'', ``a fuzzy pokemon:: by beeple and james gilleard and justin gerard :: ornate, dynamic, particulate, intricate, elegant, highly detailed, centered, artstation, smooth, sharp focus, octane render, 3''
\item \textbf{OpenWebText:} ``One family says the ratings-grabbing reality show "Extreme Makeover: Home Edition" turned their personal tragedy into a practical nightmare, leaving them with virtually nothing but a lawsuit. $\ldots$ for more analysis and interviews on the top legal stories each weeknight at 6 p.m. ET on MSNBC TV.'' (1,386 words 7,681 characters)
\item \textbf{COYO-700M:} ``Kidney humble nurse mascot design with a syringe'', ``Load image into Gallery viewer, Trevor Medium Highland Cow'', ``American Girl Doll Grace Thomas Goty Girl Of The Year 2015 + AG Outfit VGC'', ``Lajama Luxury Tempered Glass Phone Case For IPhone 6 6S 7 8 Plus Anti-scratch Silicone Protector Glass Back Cover For Iphone X''
\end{itemize}

\section{More experiments}
\label{subsec:appendix_more_experiments}

\subsection{Norm distributions of different probabilistic distributions}

\begin{figure}[t]
    \centering
    \begin{minipage}[b]{0.5\textwidth}
    \centering
        \includegraphics[width=\linewidth]{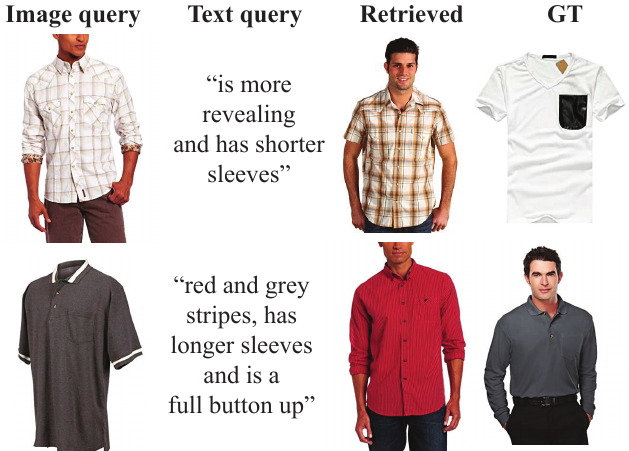} 
        \caption{\small {\bf False negatives in CIR datasets.} Examples are drawn from FashionIQ \cite{fashioniq}, retrieved by \ours.}
        \label{fig:fn}
    \end{minipage}
    \hfill
    \begin{minipage}[b]{0.45\textwidth}
    \centering
        \includegraphics[width=\linewidth]{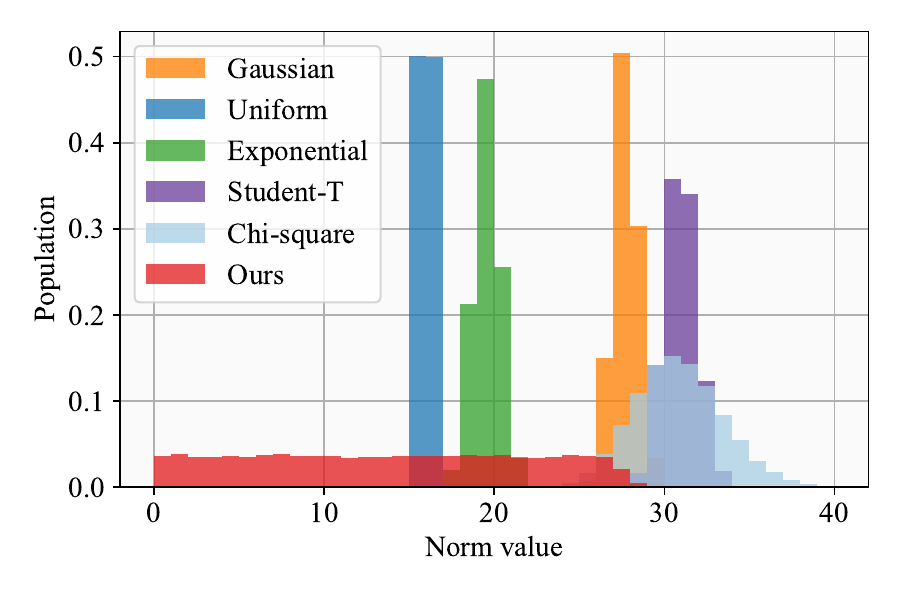} 
        \vspace{-1em}
        \caption{\small {\bf Norm distribution of different distributions.} The statistics are measured by 768-dim random vectors (the CLIP ViT-L/14 textual embedding dimension) drawn from six different probabilistic distributions. ``Ours'' denotes $\text{Unif}(0, 1) \times \mathcal N(0, 1)$.}
        \label{fig:norm_dist}
    \end{minipage}
\end{figure}

\cref{fig:norm_dist} shows the distribution of the norms of the 768-dimensional random vectors drawn from each probability distribution. Here, we observe that the samples drawn from a Gaussian distribution have almost identical norm sizes. Similarly, other probability distributions, such as uniform, exponential, $\chi^2$, and student-t distributions, suffer from the same problem.
In other words, regardless of the actual gap between an image-text pair, a random noise sampled from such distributions always adds additional information to the textual latent embedding with an almost fixed amount regarding its norm. However, in practice, the gap between image-text pairs can be diverse.
We presume this less diverse norm size of the added random noise restricts the generalizability against the modality gap.
\cref{fig:norm_dist} also illustrates that our design choice shows a more diverse norm distribution than the other distributions.

\subsection{\ours with other backbones}

\cref{tab:ours_blip} shows the results of \ours with BLIP ViT-B/16 backbone \cite{blip}. Similar to CLIP, the BLIP ViT-B/16 backbone uses ViT-B/16 \cite{vit} as the image encoder and Transformer \cite{vaswani2017attention} as the text encoder. We use the BLIP encoders as the separated feature encoders as the CLIP encoders, instead of using the joint multimodal embedding extraction. The table shows that the BLIP backbone shows a comparable performance to the CLIP backbone, despite of using a smaller backbone.

\begin{table}[h]
\small
\centering
\begin{tabular}{cccccc}
\toprule
&CIRCO & GeneCIS & FashionIQ & CIRR & \multirow{2}{*}{Avg} \\
&mAP@5 & R@3 & R@10 & R@10 & \\ \midrule
CLIP ViT-L/14 & {12.59} & {32.38} & {26.28} & {66.68} & {34.48} \\
BLIP ViT-B/16 \cite{blip} & {12.75} & {29.65} & {25.00} & {65.35} & {33.19} \\ 
\bottomrule
\end{tabular}
\caption{\small {\bf BLIP results.}}
\label{tab:ours_blip}
\end{table}

\subsection{Comparison with more methods}
In the main paper, we compare \ours with two ZS CIR methods: Pic2Word \cite{pic2word} and SEARLE \cite{searle}. In this subsection, we compare \ours with more recent CIR methods, such as CompoDiff \cite{compodiff} and CoVR \cite{ventura23covr}. The most significant drawback of CompoDiff and CoVR is a heavy computation. While \ours only needs 30 mins to train a high performing ZS CIR model, CompoDiff and CoVR take more than a day even with a ViT-L backbone.

CompoDiff tackles ZS CIR problem by generating massive 18.8M synthetic triplets and employing image feature editing strategy via a latent diffusion model. Despite its ability to handle negative text conditions and mask conditions, CompoDiff suffers from heavy computations. \cref{tab:appendix_time_comparison} shows the full comparisons of CompoDiff and the other ZS CIR methods, including \ours. We can observe that CompoDiff needs a heavy GPU computations (about 10 days with 128 A100 for training) and a relatively slow inference time (0.12 vs. 0.02).
Although CompoDiff has better flexibility than \ours, we do not directly compare CompoDiff with \ours due to its heavy resource computation.

\begin{table}[h]
\small
\centering
\begin{tabular}{lllllll}
\toprule
 &  & Training time1 (h) & Training time2 (h) & Total training time (h) & Inference time (s) & Training GPUs \\ \midrule
\multirow{4}{*}{ViT-L} & Pic2Word & 3.0 & - & 3.0 & 0.02 & A100 x 8 \\
 & SEARLE & 1.7 & 2.5 & 4.2 & 0.02 & A100 x 8 \\
 & \ours & 0.5 & - & 0.5 & 0.02 & A100 x 8 \\
 & CompoDiff & 6 days, 10hours & 3 days, 5hours & 9 days, 15hours & 0.12 & A100 x 128 \\ \midrule
\multirow{3}{*}{ViT-H} & Pic2Word & 7.3 & - & 7.3 & 0.035 & A100 x 8 \\
 & SEARLE & 3.6 & 4.5 & 8.1 & 0.042 & A100 x 8 \\
 & \ours & 0.7 & - & 0.7 & 0.042 & A100 x 8 \\ \midrule
\multirow{3}{*}{ViT-G} & Pic2Word & 13.4 & - & 13.4 & 0.050 & A100 x 8 \\
 & SEARLE & 6.3 & 8.1 & 14.4 & 0.047 & A100 x 8 \\
 & \ours & 0.8 & - & 0.8 & 0.047 & A100 x 8 \\ \bottomrule
\end{tabular}
\caption{\small {\bf Training time and inference time comparisons.} Pic2Word and \ours is trained on a single stage, while SEARLE and CompoDiff need a two stage training strategy. The inference time is measured by a single A100 GPU with a single image.}
\label{tab:appendix_time_comparison}
\end{table}

Similarly, we do not directly compare \ours with CoVR because CoVR relies on the joint multi-modal encoder of BLIP \cite{blip} and a large-scale video dataset (1.6M triplets). We focus on a fair comparison of CIR methods that share the same embedding space (\ie, the CLIP latent embedding space), without using a multimodal cross-attention Transformer taking both image and text inputs to compute a multimodal embedding. CoVR also uses a larger image resolution (384 pixels) than our comparison methods (224 pixels).

\subsection{GeneCIS full results}

\cref{tab:appendix_genecis_full} shows the full results of the comparison methods on GeneCIS. \cref{tab:appendix_genecis_avg} shows the average scores for ``Focus'', ``Change'', ``Attribute'', and ``Object'' tasks. In the table, we observe that \ours shows significantly better performances than others, especially for ``Focus'' tasks and ``Attribute'' tasks. We presume that our keyword masking strategy for \smp improves the ability to understand short keywords of GeneCIS, such as ``color'' in \cref{fig:dataset_genecis_focus_att}.

\begin{table}[t]
\small
\setlength{\tabcolsep}{4pt}
\centering
\begin{tabular}{ll|lll|lll|lll|lll|lll}
\toprule
 &  & \multicolumn{3}{c|}{Focus Attribute} & \multicolumn{3}{c|}{Change Attribute} & \multicolumn{3}{c|}{Focus Object} & \multicolumn{3}{c|}{Change Object} & \multicolumn{3}{c}{Average} \\
 &  & R@1 & R@2 & R@3 & R@1 & R@2 & R@3 & R@1 & R@2 & R@3 & R@1 & R@2 & R@3 & R@1 & R@2 & R@3 \\ \midrule
\multirow{3}{*}{ViT-L} & Pic2Word & 15.65 & 28.16 & 38.65 & 13.87 & 24.67 & 33.05 & 8.42 & 18.01 & 25.77 & 6.68 & 15.05 & 24.03 & 11.16 & 21.47 & 30.38 \\
 & SEARLE & 17.00 & 29.65 & 40.70 & 16.38 & 25.28 & 34.14 & 7.76 & 16.68 & 25.31 & 7.91 & 16.84 & 25.05 & \best{12.26} & 22.11 & 31.30 \\
 & \ours & 16.90 & 29.95 & 41.45 & 16.19 & 27.98 & 36.84 & 8.27 & 17.40 & 26.22 & 7.40 & 15.71 & 25.00 & 12.19 & \best{22.76} & \best{32.38} \\ \midrule
\multirow{3}{*}{ViT-H} & Pic2Word & 18.60 & 30.70 & 42.10 & 13.16 & 23.91 & 33.14 & 9.23 & 17.60 & 27.14 & 6.58 & 16.48 & 25.36 & 11.89 & 22.17 & 31.94 \\
 & SEARLE & 18.75 & 31.50 & 42.25 & 15.53 & 26.85 & 35.89 & 10.61 & 18.67 & 26.53 & 8.47 & 17.86 & 26.22 & 13.34 & 23.72 & 32.72 \\
 & \ours & 19.60 & 31.50 & 41.55 & 16.62 & 27.60 & 37.50 & 9.80 & 18.83 & 27.86 & 9.03 & 17.55 & 25.71 & \best{13.76} & \best{23.87} & \best{33.16} \\ \midrule
\multirow{3}{*}{ViT-G} & Pic2Word & 12.45 & 23.40 & 33.65 & 11.74 & 21.88 & 30.87 & 9.90 & 19.34 & 27.35 & 8.57 & 18.16 & 26.12 & 10.67 & 20.70 & 29.50 \\
 & SEARLE & 16.30 & 29.40 & 40.70 & 16.15 & 27.32 & 35.46 & 10.77 & 18.16 & 27.91 & 8.27 & 15.56 & 25.77 & 12.87 & 22.61 & 32.46 \\
 & \ours & 19.05 & 33.00 & 42.30 & 17.57 & 30.16 & 38.07 & 10.10 & 19.08 & 28.06 & 7.91 & 16.33 & 25.71 & \best{13.66} & \best{24.64} & \best{33.54} \\
 \bottomrule
\end{tabular}
\vspace{-.5em}
\caption{\small {\bf GeneCIS full results.}}
\label{tab:appendix_genecis_full}
\end{table}
\vspace{-.5em}
\begin{table}[t]
\small
\setlength{\tabcolsep}{4pt}
\centering
\begin{tabular}{ll|lll|lll|lll|lll}
\toprule
 &  & \multicolumn{3}{c|}{``Focus'' Avg} & \multicolumn{3}{c|}{``Change'' Avg} & \multicolumn{3}{c|}{``Attribute'' Avg} & \multicolumn{3}{c}{``Object'' Avg}\\
 &  & R@1 & R@2 & R@3 & R@1 & R@2 & R@3 & R@1 & R@2 & R@3 & R@1 & R@2 & R@3 \\ \midrule
\multirow{3}{*}{ViT-L} & Pic2Word & 14.76 & 26.42 & 35.85 & 7.55 & 16.53 & 24.90 & 12.04 & 23.09 & 32.21 & 10.28 & 19.86 & 28.54 \\
 & SEARLE & \best{16.69} & 27.47 & 37.42 & \best{7.84} & \best{16.76} & 25.18 & 12.38 & 23.17 & 33.01 & \best{12.15} & 21.06 & 29.60 \\
 & \ours & 16.55 & \best{28.97} & \best{39.15} & \best{7.84} & 16.56 & \best{25.61} & \best{12.59} & \best{23.68} & \best{33.84} & 11.80 & \best{21.85} & \best{30.92} \\ \midrule
\multirow{3}{*}{ViT-H} & Pic2Word & 15.88 & 27.31 & 37.62 & 7.91 & 17.04 & 26.25 & 13.92 & 24.15 & 34.62 & 9.87 & 20.20 & 29.25 \\
 & SEARLE & 17.14 & 29.18 & 39.07 & \best{9.54} & \best{18.27} & 26.38 & 14.68 & 25.09 & 34.39 & 12.00 & 22.36 & 31.06 \\
 & \ours & \best{18.11} & \best{29.55} & \best{39.53} & 9.42 & 18.19 & \best{26.79} & \best{14.70} & \best{25.17} & \best{34.71} & \best{12.83} & \best{22.58} & \best{31.61} \\ \midrule
\multirow{3}{*}{ViT-G} & Pic2Word & 12.10 & 22.64 & 32.26 & 9.24 & \best{18.75} & 26.74 & 11.18 & 21.37 & 30.50 & 10.16 & 20.02 & 28.50 \\
 & SEARLE & 16.23 & 28.36 & 38.08 & \best{9.52} & 16.86 & 26.84 & 13.54 & 23.78 & 34.31 & 12.21 & 21.44 & 30.62 \\
 & \ours & \best{18.31} & \best{31.58} & \best{40.19} & 9.01 & 17.71 & \best{26.89} & \best{14.58} & \best{26.04} & \best{35.18} & \best{12.74} & \best{23.25} & \best{31.89} \\
 \bottomrule
\end{tabular}
\vspace{-.5em}
\caption{\small {\bf GeneCIS average results.}}
\label{tab:appendix_genecis_avg}
\vspace{-.5em}
\end{table}

\subsection{Exploring inference prompt}
We also explore the impact of the ZS CIR prompt rather than ``\aphotopromptwithcond''. \cref{tab:appendix_prompts} shows the results. In the paper, we observe that choosing a different prompt can significantly enhance retrieval performance. For example, by changing ``\aphotopromptwithcond'' to ``Observe \sptkn that \condtkn'', \ours R@50 is improved by almost 2.0pp (46.48 to 48.41). On the other hand, we observe that Pic2Word is rarely improved by changing the prompt. SEARLE shows a reasonable improvement compared to Pic2Word but still performs worse than \ours.

\begin{table}[ht!]
\footnotesize
\setlength{\tabcolsep}{3.5pt}
\centering
\begin{tabular}{lcccccc}
\toprule
 & Pic2Word R@10 & SEARLE R@10 & \ours R@10 & Pic2Word R@50 & SEARLE R@50 & \ours R@50 \\ \midrule
\rowcolor{secondcolor}
a photo of \sptkn that \condtkn & \best{25.51} & 24.64 & 26.28 & 44.98 & 44.41 & 46.48 \\
\sptkn that \condtkn & 24.83 & 25.05 & 27.00 & 45.09 & 44.90 & 47.56 \\
\sptkn with \condtkn & 24.16 & 25.08 & 26.99 & 43.03 & 44.77 & 47.62 \\
\sptkn , \condtkn & 25.20 & 24.39 & 27.08 & 43.93 & 44.40 & 47.87 \\
\sptkn adapted to \condtkn & 23.31 & 24.92 & 26.36 & 42.82 & 43.97 & 47.63 \\
\sptkn modified by \condtkn & 23.56 & 23.83 & 26.02 & 42.37 & 43.20 & 46.29 \\
\sptkn in response to \condtkn & 23.31 & 25.10 & 26.71 & 41.90 & 44.82 & 47.31 \\
\sptkn transformed by \condtkn & 24.02 & 24.10 & 26.45 & 42.81 & 43.69 & 46.52 \\
\sptkn influenced by \condtkn & 21.30 & 23.52 & 26.52 & 39.76 & 43.16 & 47.28 \\
Retrieval of \sptkn using feedback \condtkn & 21.01 & 22.13 & 24.82 & 38.98 & 40.99 & 44.96 \\
\sptkn guided by \condtkn & 24.01 & 24.40 & 26.56 & 42.84 & 44.52 & 47.15 \\
\sptkn adjusted to \condtkn & 24.05 & 24.63 & 26.78 & 43.68 & 44.55 & 47.67 \\
\sptkn in alignment with \condtkn & 22.49 & 23.93 & 26.07 & 40.43 & 42.19 & 46.38 \\
\sptkn in correspondence to \condtkn & 22.55 & 23.60 & 26.44 & 41.23 & 42.39 & 46.55 \\
\sptkn refined with \condtkn & 22.91 & 23.58 & 26.66 & 41.59 & 43.17 & 46.89 \\
\sptkn as directed by \condtkn & 23.28 & 25.59 & 26.96 & 42.17 & 45.30 & 47.70 \\
\sptkn evolved from \condtkn & 24.64 & 23.12 & 26.77 & 44.02 & 42.63 & 47.26 \\
\sptkn inspired by \condtkn & 24.35 & 24.18 & 26.26 & 43.71 & 43.61 & 47.05 \\
\sptkn with adjustments from \condtkn & 23.76 & 24.11 & 26.26 & 42.11 & 44.64 & 46.92 \\
\sptkn in consideration of \condtkn & 22.20 & 23.38 & 26.78 & 40.87 & 42.26 & 47.11 \\
\sptkn , taking into account \condtkn & 23.71 & 24.18 & 26.44 & 42.74 & 43.08 & 47.23 \\
\sptkn as influenced by the query \condtkn & 21.45 & 22.92 & 25.86 & 39.72 & 43.04 & 45.72 \\
\sptkn reshaped by \condtkn & 24.05 & 24.31 & 26.07 & 43.17 & 43.68 & 46.25 \\
\sptkn curated based on \condtkn & 25.05 & 24.86 & 26.48 & \best{45.36} & 44.72 & 47.11 \\
\sptkn showcasing \condtkn & 23.68 & 24.27 & 26.50 & 42.68 & 44.48 & 46.80 \\
An instance of \sptkn where \condtkn & 23.32 & 24.50 & 25.71 & 42.09 & 43.24 & 45.84 \\
\sptkn highlighting \condtkn & 24.45 & 22.34 & 26.19 & 42.98 & 41.58 & 46.40 \\
A depiction of \sptkn exhibiting \condtkn & 22.26 & 23.42 & 25.53 & 41.97 & 42.79 & 45.27 \\
\sptkn as exemplified by \condtkn & 23.58 & 24.95 & 26.35 & 42.98 & 44.89 & 46.78 \\
\sptkn demonstrating \condtkn & 24.21 & 24.59 & 25.78 & 43.24 & 44.32 & 45.89 \\
An illustration of \sptkn portraying \condtkn & 24.71 & 25.15 & 26.18 & 43.95 & 44.70 & 46.47 \\
\sptkn in the context of \condtkn & 24.08 & 25.07 & 27.16 & 43.56 & 44.31 & 47.56 \\
\sptkn as influenced by \condtkn & 21.30 & 24.37 & 26.61 & 40.22 & 43.97 & 46.89 \\
\sptkn characterized by \condtkn & 24.21 & 23.40 & 26.64 & 44.04 & 42.28 & 46.49 \\
\sptkn : An exploration of \condtkn & 23.51 & 25.52 & 26.25 & 42.66 & 45.45 & 46.59 \\
A presentation of \sptkn underlined by \condtkn & 22.93 & 22.48 & 24.86 & 40.44 & 41.09 & 44.03 \\
A manifestation of \sptkn reflecting \condtkn & 22.55 & 22.71 & 25.25 & 40.58 & 42.04 & 45.39 \\
\sptkn in light of \condtkn & 22.62 & 25.03 & 26.41 & 41.70 & 44.71 & 46.99 \\
\sptkn as a testament to \condtkn & 22.02 & 24.25 & 26.93 & 41.28 & 44.19 & 47.65 \\
\sptkn intertwined with \condtkn & 24.45 & 23.64 & 25.06 & 43.92 & 42.37 & 45.15 \\
\sptkn complemented by \condtkn & 25.23 & 24.21 & 26.32 & 45.04 & 43.23 & 47.07 \\
\sptkn juxtaposed with \condtkn & 25.40 & 24.51 & 26.76 & 44.73 & 43.10 & 47.43 \\
A representation of \sptkn in relation to \condtkn & 23.26 & 23.83 & 25.56 & 41.69 & 43.31 & 46.31 \\
\sptkn that \condtkn & 24.83 & 25.05 & 27.00 & 45.09 & 44.90 & 47.56 \\
\sptkn which \condtkn & 24.29 & 24.58 & 27.07 & 43.62 & 44.69 & 47.74 \\
\sptkn where it \condtkn & 24.62 & 25.20 & 27.12 & 44.30 & 44.65 & 47.97 \\
Discover \sptkn that \condtkn & 25.39 & 25.12 & 26.07 & 45.02 & 44.82 & 46.24 \\
Retrieve \sptkn that \condtkn & 23.86 & 24.47 & 26.61 & 42.60 & 43.78 & 46.64 \\
Search for \sptkn that \condtkn & 24.39 & 24.04 & 26.83 & 44.15 & 43.50 & 47.17 \\
Identify \sptkn which \condtkn & 22.49 & 23.67 & 26.46 & 41.35 & 43.60 & 46.70 \\
Highlight \sptkn that \condtkn & 24.53 & 23.45 & 26.01 & 43.89 & 42.72 & 46.65 \\
Present \sptkn where it \condtkn & 24.12 & 24.40 & 27.29 & 43.58 & 43.66 & 47.94 \\
Showcase \sptkn that \condtkn & 24.63 & 24.68 & 26.63 & 44.00 & 44.42 & 47.26 \\
Explore \sptkn which \condtkn & 24.46 & 23.72 & 26.40 & 42.88 & 43.79 & 46.02 \\
Find \sptkn that \condtkn & 24.65 & 25.03 & 27.28 & 44.97 & 44.44 & 47.73 \\
Source \sptkn which \condtkn & 24.67 & 24.38 & 27.16 & 44.24 & 44.03 & 47.63 \\
View \sptkn where it \condtkn & 24.06 & 24.96 & 27.16 & 43.56 & 43.93 & 47.99 \\
Examine \sptkn that \condtkn & 25.10 & 24.53 & 26.64 & 44.89 & 43.94 & 47.65 \\
Analyze \sptkn which \condtkn & 23.90 & 23.66 & 26.05 & 42.32 & 42.92 & 46.27 \\
Observe \sptkn that \condtkn & 24.53 & \best{26.50} & \best{27.42} & 44.78 & \best{45.83} & \best{48.41} \\
Report \sptkn which \condtkn & 23.91 & 23.69 & 26.48 & 42.92 & 43.11 & 47.04 \\
See \sptkn where it \condtkn & 25.47 & 25.28 & 27.24 & 44.76 & 45.14 & 48.24 \\
Document \sptkn that \condtkn & 24.49 & 24.52 & 26.44 & 44.31 & 44.58 & 47.17 \\ \midrule
Average performance & 23.82 & 24.27 & 26.44 & 42.96 & 43.76 & 46.88 \\
Best prompt & 25.51 & 26.50 & \best{27.42} & 45.36 & 45.83 & \best{48.41} \\ \bottomrule
\end{tabular}
\vspace{-1em}
\caption{\small {\bf FashionIQ R@10 and R@50 by varying text prompts.}}
\label{tab:appendix_prompts}
\end{table}

\section{Qualitative results}

\subsection{Retrieval from LAION}

\begin{figure}
    \centering
    \includegraphics[width=\linewidth]{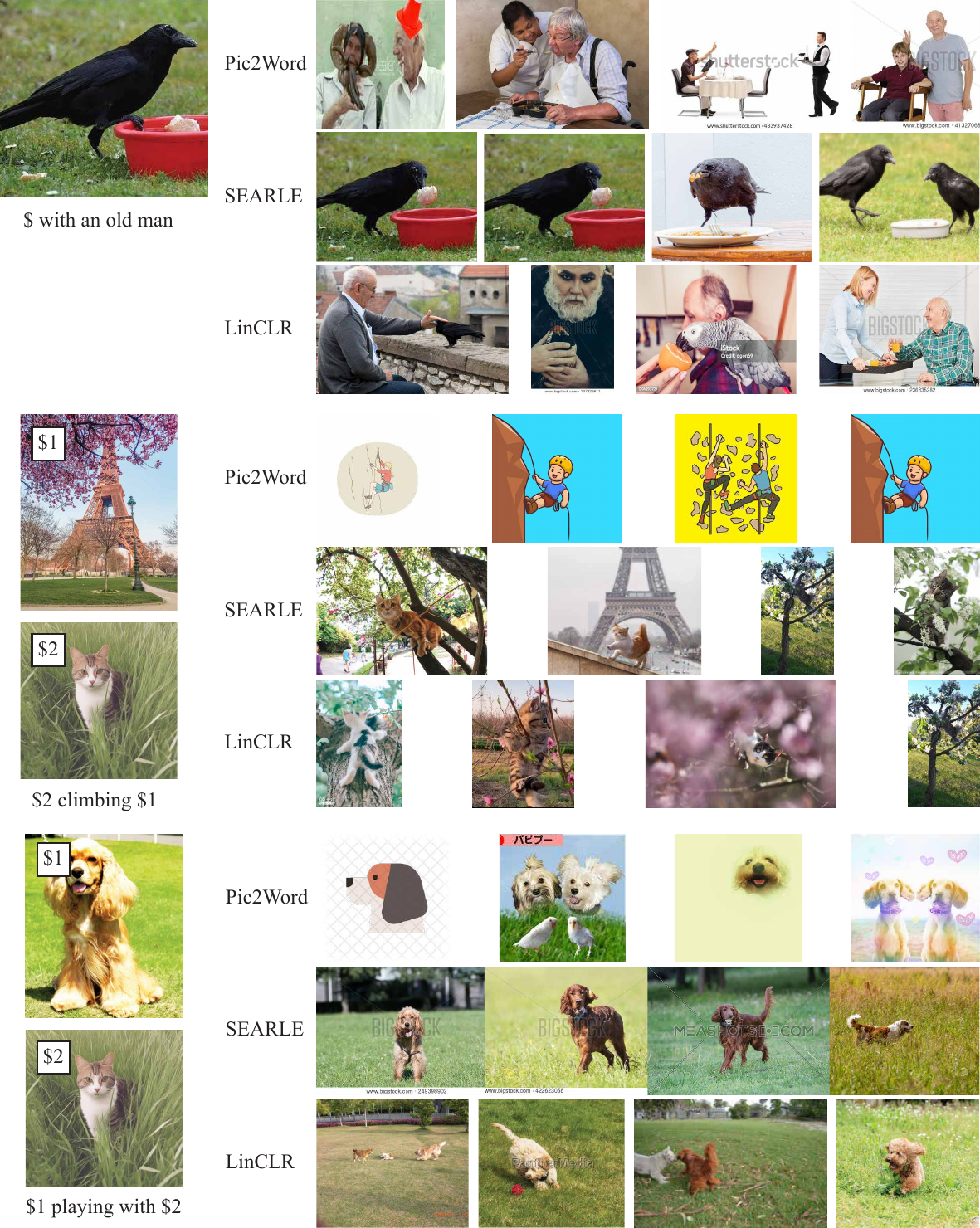}
    \caption{\small {\bf Retrieval results from LAION-2B CLIP-L features.} More examples are in \texttt{more\_examples.pdf}.}
    \label{fig:appendix_laion_retrieval}
\end{figure}

We compare qualitative retrieval results of the comparison methods on a million-scale search database using LAION-2B. 
\cref{fig:appendix_laion_retrieval} shows the results using CLIP ViT-L features. We include more examples in \url{https://github.com/navervision/lincir}, including ViT-H features.
In the figure, we observe that Pic2Word cannot handle both image and text conditions. For example, in the first example, Pic2Word ignores the crow visual information but only focuses on the ``old man'' text query. Similarly, the second example shows that Pic2Word ignores the visual information from the Eiffel Tower and the cat images but only focuses on climbing. SEARLE shows better results than Pic2Word, but as shown in the first example, SEARLE often attends to the visual information rather than the textual query.
Interestingly, although our method is not trained on multiple query examples as the second and third examples, it shows reasonable retrieval results.

\clearpage
{
    \small
    \bibliographystyle{ieeenat_fullname}
    \bibliography{main}
}

\end{document}